\pdfoutput=1
\documentclass[10pt]{article}
\usepackage{etoolbox}
\PassOptionsToPackage{table}{xcolor}
\usepackage{leapwing}
\hypersetup{colorlinks=true,allcolors=LWInk}
\usepackage{amsmath,amsfonts,bm}

\def\eqref#1{equation~\ref{#1}}
\def\1{\bm{1}}

\DeclareMathAlphabet{\mathsfit}{\encodingdefault}{\sfdefault}{m}{sl}
\SetMathAlphabet{\mathsfit}{bold}{\encodingdefault}{\sfdefault}{bx}{n}

\usepackage{wrapfig}
\usepackage{subcaption}
\usepackage{multirow}
\definecolor{referencegray}{gray}{0.94}

\renewcommand{\LWTitle}{PoseVLN: Learning Geometric State Evolution for Vision-Language Navigation}
\renewcommand{\LWShortTitle}{PoseVLN}
\renewcommand{\LWReportID}{Technical Report}
\renewcommand{\LWDate}{November 2026}
\renewcommand{\LWPartnerLogoHeight}{11mm}
\renewcommand{\LWPartnerLogos}{%
  \LWPartnerLogo[12mm]{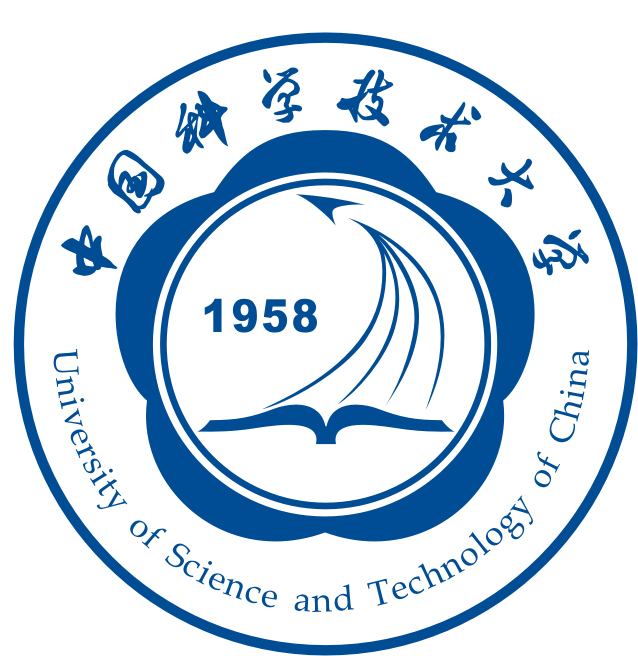}\hspace{3mm}%
  \LWPartnerLogo[12mm]{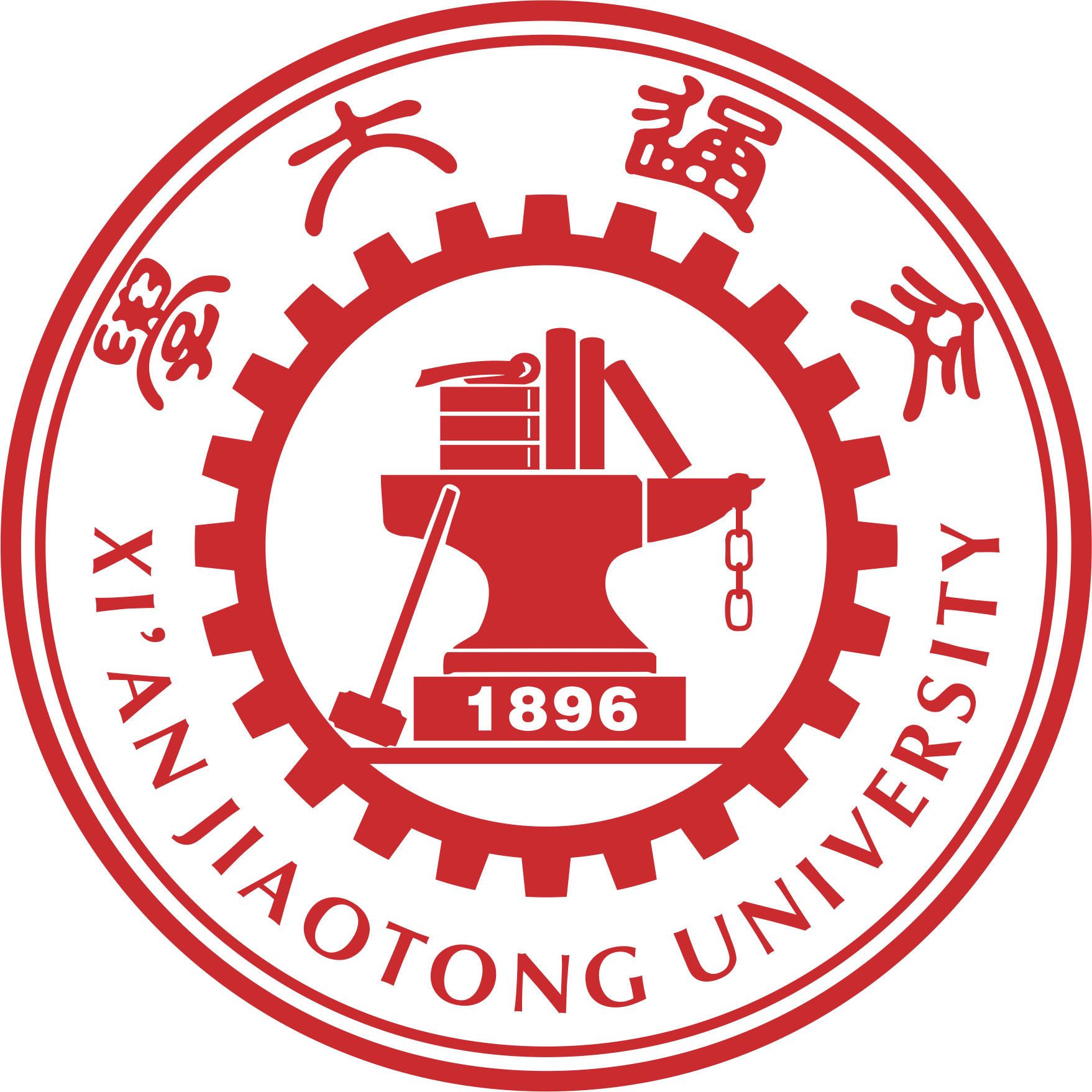}%
}
\renewcommand{\LWAuthors}{%
  Pengyu Hou\textsuperscript{1*}\quad
  Mingfei Han\textsuperscript{2,3,4*\textdagger}\quad
  Haihong Hao\textsuperscript{3}\quad
  Xiaolong Wu\textsuperscript{3}\\[3pt]
  Bo Dong\textsuperscript{1}\quad
  Xiaojun Chang\textsuperscript{2,3}\quad
  Salman Khan\textsuperscript{4}\quad
  Bin Shi\textsuperscript{1}}
\renewcommand{\LWAffiliations}{%
  \textsuperscript{1}Xi'an Jiaotong University~
  \textsuperscript{2}LeapWing AI~
  \textsuperscript{3}Future Media Computing Lab, USTC~
  \textsuperscript{4}MBZUAI}
\renewcommand{\LWNotes}{%
  {\sffamily\color{LWMuted}
  \textsuperscript{*}Equal contribution\quad
  \textsuperscript{\textdagger}Project lead\quad
    }
}
\renewcommand{\LWLinks}{%
  \textbf{Project page:} {\hypersetup{urlcolor=LWIndigo}\url{https://houpengyu123.github.io/PoseVLN/}}%
}
\renewcommand{\LWAbstract}{%
  \begingroup
  \renewenvironment{abstract}{}{}%
  \begin{abstract}
Vision-Language Navigation (VLN) requires an embodied agent to relate past experience, current spatial state, and future motion while following natural-language instructions.
Recent approaches improve these capabilities through long-horizon memory, state-aware reasoning, and future prediction, yet these components are often modeled separately, leaving navigation state evolution insufficiently connected across time.
We argue that camera pose provides a natural geometric representation for bridging past, present, and future states.
Based on this insight, we propose PoseVLN, a pose-aware VLN framework that jointly learns navigation decisions and geometric state evolution.
PoseVLN organizes visual memory and the current observation in an episode-relative coordinate frame, and jointly predicts the next $K$ navigation actions together with a local $\Delta$Pose over the same horizon.
The predicted $\Delta$Pose is composed with the current trajectory-level pose to obtain the future pose state, which is recursively propagated throughout navigation, forming a continuous geometric state evolution from past memory to future motion.
Experiments on the R2R-CE and RxR-CE Val-Unseen splits demonstrate the effectiveness of explicitly modeling geometric state evolution, with PoseVLN outperforming the strongest compared baselines by 2.6 and 2.9 percentage points in success rate, respectively, using only single RGB observations.

\end{abstract}

  \endgroup
}

\begin{document}
\makeLWTitle

\section{Introduction}
\label{sec:introduction}
Vision-Language Navigation (VLN) requires an embodied agent to follow natural-language instructions and navigate to a target by interpreting visual observations and taking appropriate actions. Recent advances have extended VLN beyond direct instruction and observation matching, with increasing attention to memory, spatial reasoning, and predictive modeling. Memory-based approaches preserve information from previous observations and decisions~\citep{wei2025streamvln,song2025towards,zhang2025mapnav,xue2026profocus,liu2026memvln}, while state-aware and geometric methods support navigation through progress reasoning, structured scene representations, or explicit spatial features~\citep{wang2026progress,wang2026dynam3d,li2026bridging,zhang2025mapnav,xue2026profocus}. Predictive approaches further anticipate future actions, observations, landmarks, or latent dynamics before execution~\citep{yao2025navmorph,dai2026history,huang2026wnm,xu2025navq,wang2026monodream,wang2026think,zhang2026futurenav}. Together, these directions highlight the importance of relating past experience, the current navigation context, and future motion.

Despite these advances, two aspects of spatial modeling remain insufficiently addressed. First, past and current observations are not necessarily anchored in an explicit common spatial reference. Retrieving a relevant memory therefore does not by itself establish its viewpoint's position and orientation relative to the agent, leaving this relationship to be inferred for spatial decision-making. Second, future prediction does not necessarily specify an explicit geometric transition of the navigation state. Anticipated actions or visual content can guide action selection without providing an updated position and orientation for subsequent decisions. These gaps motivate a common geometric representation that explicitly relates past observations to the current state and describes how that state changes over the upcoming action horizon.

Camera pose provides a compact representation for addressing both needs. A trajectory-level pose specifies the position and orientation of the camera in a shared coordinate frame, associating each past and current observation with the spatial state from which it was acquired. Relative poses between these observations then make their spatial relationships explicit. For future motion, a local pose change describes the translation and rotation over the upcoming action horizon. Composing this change with the current pose yields a future pose in the same coordinate frame, allowing the resulting state to be carried forward and associated with subsequent observations. Together, trajectory-level poses and local pose changes provide a common geometric description of past observations, the current viewpoint, and future motion. We refer to this continual maintenance and update of the agent's spatial state as {geometric state evolution}.

\begin{figure}[t]
    \centering
    \resizebox{\linewidth}{!}{%
        \includegraphics[height=5cm,keepaspectratio]{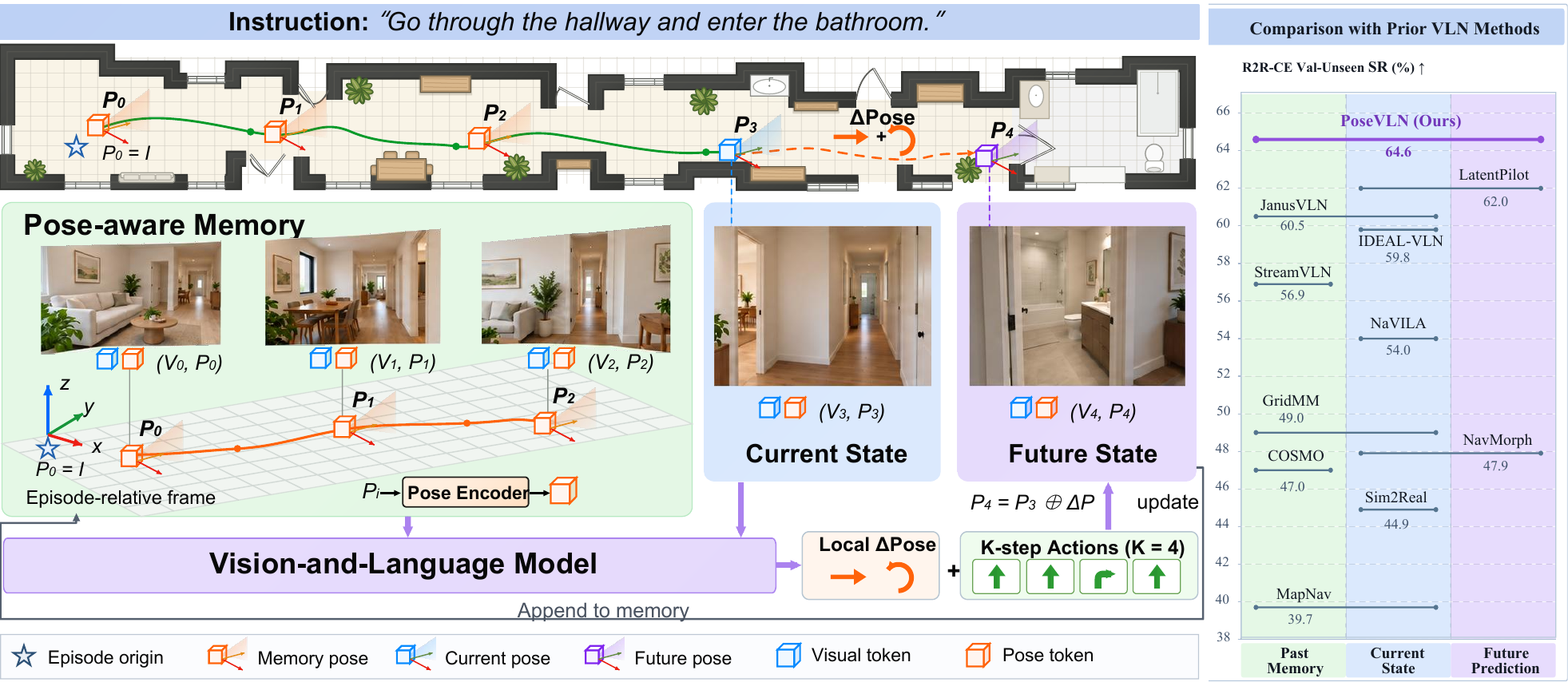}%
    }
    \caption{
The PoseVLN framework and comparison with prior methods. Left: pose-aware memory and geometric state evolution.
Right: comparison with prior VLN methods by navigation performance and temporal focus.
}
    \label{fig:intro}
    \vspace{-6pt}
\end{figure}

Based on this insight, we introduce {PoseVLN}, a pose-aware VLN framework that explicitly models navigation as geometric state evolution in a trajectory-level coordinate space. We define the first observation as the origin of an episode-relative coordinate system and associate each subsequent visual observation in memory with a trajectory-level pose state, allowing past observations and the current state to be organized within a common spatial reference. As illustrated in Figure~\ref{fig:intro}, given the instruction, current observation, visual memory, and their associated pose states, PoseVLN jointly predicts the next $K$ navigation actions and a local $\Delta$Pose that describes the geometric state transition over the same action horizon. The predicted $\Delta$Pose is composed with the current trajectory-level pose to obtain the future pose state, which is stored back into memory and reused in subsequent navigation. During training, ground-truth trajectory poses provide supervision for geometric state transitions.
At inference, PoseVLN operates solely on a monocular RGB stream, without depth or external odometry.
The first observation is assigned the identity pose $P_0=I$ to define the episode-relative frame.
No ground-truth pose is used, and all subsequent pose states are recursively predicted by the model.
% \begin{figure}[t]
%     \centering
%     \includegraphics[width=\linewidth]{image/draw (2).pdf}
%     \caption{Overview of the key idea of PoseVLN. Camera pose provides a unified geometric representation that bridges past observations, the current spatial state, and future navigation decisions.}
%     \label{fig:intro}
% \end{figure}
Our main contributions are three-fold:
\begin{itemize}
    \item We propose PoseVLN, a pose-aware VLN framework that models
    geometric state evolution through joint action and local $\Delta$Pose
    prediction, recursively propagating poses without step-wise
    ground-truth pose or odometry at inference.

    \item We demonstrate strong performance on R2R-CE and RxR-CE,
    showing that geometric state evolution improves long-horizon
    navigation using only monocular RGB observations.

    \item We conduct comprehensive ablations and pose evaluations to analyze the relationship between pose estimation and navigation, providing new insights into their interplay.
\end{itemize}
\section{Related Work}
\label{sec:related_work}

\subsection{Memory and State-Aware Navigation}

Video-based approaches such as NaVid~\citep{zhang2024navid} and
Uni-NaVid~\citep{zhang2024uni} demonstrate the effectiveness of
VLM/VLA models for continuous navigation.
To support longer horizons, StreamVLN~\citep{wei2025streamvln}
combines fast short-term context with slow-updating compressed
memory, while Towards Long-Horizon VLN~\citep{song2025towards}
introduces multi-granularity dynamic memory.
MapNav~\citep{zhang2025mapnav}, ProFocus~\citep{xue2026profocus},
MemVLN~\citep{liu2026memvln}, and JanusVLN~\citep{zeng2026janusvln}
further improve memory utilization through semantic maps,
selective reasoning, episodic/procedural memory, or decoupled
semantic--spatial memory.
Complementarily, Dynam3D~\citep{wang2026dynam3d} constructs
hierarchical 3D representations from posed RGB-D observations,
GA-VLN~\citep{yang2026ga} uses depth and camera poses to build
geometry-aware bird's-eye-view representations, and
Progress-Think~\citep{wang2026progress} models semantic instruction
progress from observation memory.
Beyond VLN, Cambrian-P~\citep{yang2026cambrian} highlights camera
pose as a geometric signal for relating observations across
viewpoints.

\subsection{Predictive and World-Model-Based Navigation}

Predictive approaches anticipate future outcomes to support
navigation decisions.
NavMorph~\citep{yao2025navmorph} models latent environmental
dynamics for foresighted planning, while NavQ~\citep{xu2025navq}
estimates the future prospects of candidate actions.
MonoDream~\citep{wang2026monodream} predicts current and future
panoramic cues from monocular observations, while
SpaAct~\citep{li2026spaact} learns action-conditioned visual
transitions through future-frame selection.
WNM-3D~\citep{huang2026wnm} further couples future-view prediction
with navigation actions under geometry-aware scene representations.
LatentPilot~\citep{hao2026latentpilot} incorporates latent visual
reasoning to anticipate future scene information for navigation.
These methods support navigation through future prediction and
structured scene representations.
PoseVLN focuses on explicitly propagating a camera-pose state:
trajectory-level poses associate past and current observations
within a shared coordinate frame, while local $\Delta$Pose
specifies the state transition over the predicted action horizon.

\section{Method}
\label{sec:method}

\begin{figure}[t]
    \centering
    \includegraphics[width=\linewidth]{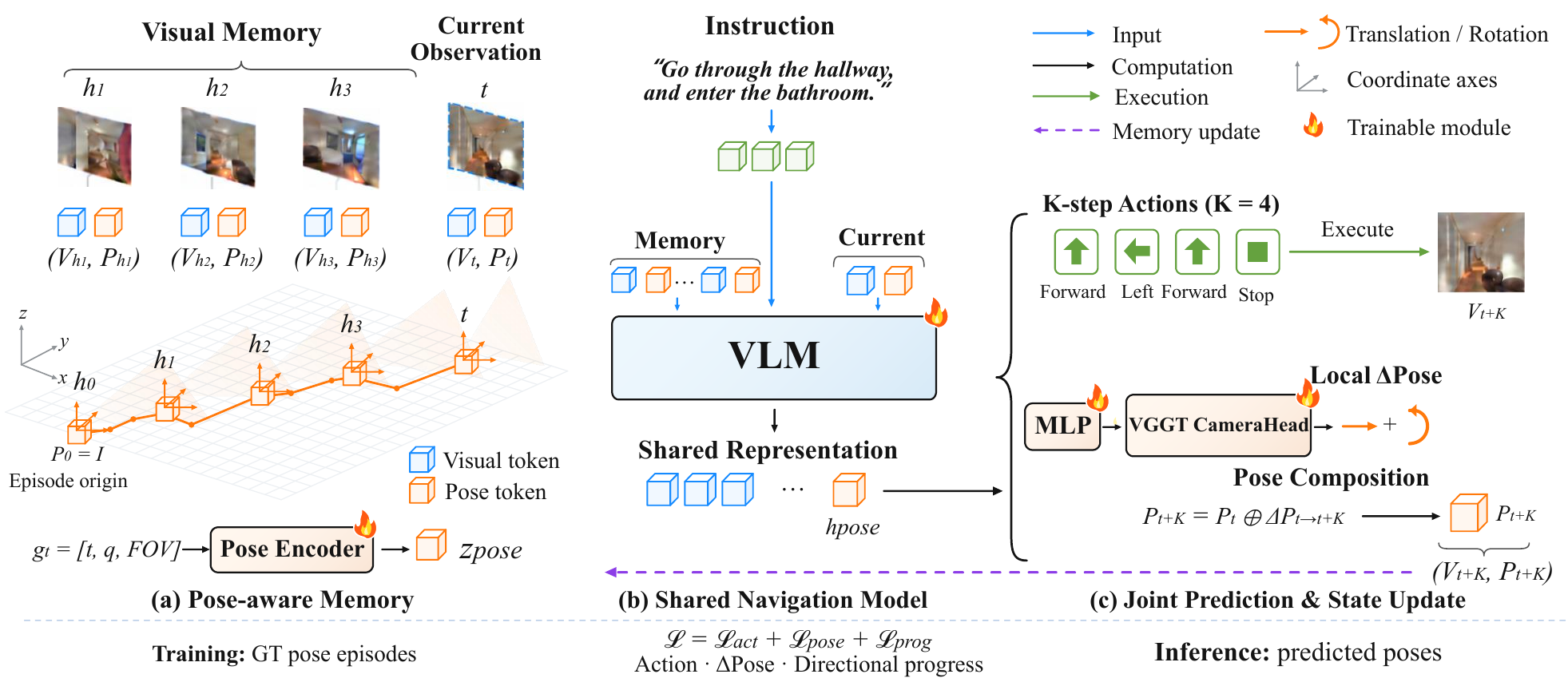}
    \caption{
    Overview of PoseVLN. Pose-aware memory associates observations with
    trajectory-level poses. A shared multimodal representation jointly predicts
    $K$-step actions and a local $\Delta$Pose, which is recursively propagated
    to update the geometric state.
    }
    \label{fig:framework}
    \vspace{-6pt}
\end{figure}

\paragraph{Framework Overview.}
We propose \textbf{PoseVLN}, a pose-aware VLN framework built upon
StreamVLN under the single-RGB streaming setting.
As illustrated in Figure~\ref{fig:framework}, pose-aware memory
associates observations with camera poses in a shared episode-relative
coordinate frame, where poses are encoded into embeddings and inserted
alongside the corresponding visual tokens.
Conditioned on these inputs and the instruction, the VLM produces
a shared representation for jointly predicting the next $K$
navigation actions and a local $\Delta$Pose over the same horizon.
The predicted transition is composed with the current trajectory-level
pose, and the resulting future pose is paired with the newly acquired
observation and written back into memory.
At inference, this process proceeds recursively from $P_0=I_4$,
requiring neither step-wise ground-truth poses nor odometry.
% \paragraph{Framework Overview.}
% We propose {PoseVLN}, a pose-aware VLN framework built upon StreamVLN
% under the single-RGB streaming setting.
% As illustrated in Figure~\ref{fig:framework}, given the instruction,
% pose-aware memory, and current observation, PoseVLN jointly predicts the
% next $K$ navigation actions and a local $\Delta$Pose.
% The predicted transition is composed with the current trajectory-level pose
% to obtain the future geometric state, which is recursively propagated and
% written back into memory for subsequent navigation.

% ============================================================
\subsection{Preliminary}
\label{sec:preliminary}

\paragraph{Navigation Task.}
Given an instruction $I$, the agent navigates from egocentric RGB observations
$\{V_t\}_{t=0}^{T}$.
At step $t$, it observes $V_t$ and maintains pose-aware memory
\begin{equation}
    \mathcal{M}_t
    =
    \{(V_{h_i},P_{h_i})\}_{i=1}^{N_h},
    \label{eq:memory}
\end{equation}
where $P_{h_i}$ denotes the trajectory-level pose associated with memory
observation $V_{h_i}$.
The action space is
$\mathcal{A}=\{\texttt{Forward},\texttt{Left},\texttt{Right},\texttt{Stop}\}$,
and the model predicts a $K$-step action chunk
\begin{equation}
    A_{t:t+K-1}
    =
    (a_t,\ldots,a_{t+K-1}),
    \qquad a_{t+k}\in\mathcal{A},
    \label{eq:action_chunk}
\end{equation}
with $K=4$ in our experiments.

\paragraph{Camera Pose Tokens.}
Let $P_i\in SE(3)$ denote the camera-to-episode transformation of $V_i$,
with $P_0=I_4$.
For tokenization, we encode its inverse $E_i^{\mathrm{w2c}}=P_i^{-1}$:
\begin{equation}
    g_i
    =
    [\mathbf t_i,\mathbf q_i,\bar f^h,\bar f^w]
    \in\mathbb{R}^{9},
    \label{eq:camera_representation}
\end{equation}
where $\mathbf t_i$ and $\mathbf q_i$ are the extrinsic translation and
unit WXYZ rotation quaternion of $E_i^{\mathrm{w2c}}$.
The FoV entries are fixed sensor constants.
A pose encoder maps $g_i$ to
$z_i^{\mathrm{pose}}=\mathrm{PoseEncoder}(g_i)$, which is concatenated
with the corresponding visual tokens.
Coordinate and parameterization details appear in
Appendices~\ref{app:episode_coordinates}
and~\ref{app:camera_parameterization}.

% ============================================================
\subsection{PoseVLN}
\label{sec:PoseVLN}

\paragraph{Joint Action and Pose Prediction.}
The instruction $I$, pose-aware memory $\mathcal{M}_t$, and current
observation--pose pair $(V_t,P_t)$ are jointly processed by the VLM
to predict navigation actions and a local pose transition over
the same $K$-step horizon:
\begin{equation}
    \left(
        I,\mathcal{M}_t,(V_t,P_t)
    \right)
    \longrightarrow
    \left(
        A_{t:t+K-1},
        \Delta\hat P_{t\rightarrow t+K}
    \right).
    \label{eq:joint_prediction}
\end{equation}
The shared representation generates the $K$ action tokens, while
the current pose-token hidden state passes through an MLP and a
VGGT CameraHead~\citep{wang2025vggt} for iterative local-pose regression.
The predicted transition is then decoded and composed with the
current pose to update the geometric state.

\paragraph{Pose-aware Geometric State Update.}
All memory and current poses share the same episode-relative frame.
Using column-vector coordinates, the local transition is
$\Delta P_{t\rightarrow t+K}=P_t^{-1}P_{t+K}$.
The CameraHead parameterizes this transition through its inverse
w2c encoding,
$\Delta E_{t\rightarrow t+K}^{\mathrm{w2c}}
=(\Delta P_{t\rightarrow t+K})^{-1}$.
During decoding, raw quaternion outputs are normalized and
converted to rotation matrices.
The future trajectory-level pose is then obtained by
\begin{equation}
    \hat P_{t+K}
    =
    P_t \oplus \Delta\hat P_{t\rightarrow t+K}
    =
    P_t
    \left(\Delta\hat E_{t\rightarrow t+K}^{\mathrm{w2c}}\right)^{-1},
    \label{eq:pose_propagation}
\end{equation}
where $\oplus$ denotes rigid-body pose composition.
After the action chunk is executed and $V_{t+K}$ becomes available,
the pair $(V_{t+K},\hat P_{t+K})$ is written back into memory
for subsequent predictions.
At inference, this update is applied recursively from $P_0=I_4$,
with $P_t$ given by the previously predicted state.
Local-transform construction and recursive propagation are detailed
in Appendices~\ref{app:delta_pose}
and~\ref{app:pose_propagation}.

% ============================================================
\subsection{Training Objective}
\label{sec:training}

PoseVLN is trained with action prediction, pose regression, and directional
progress objectives.

\paragraph{Action Loss.}
The predicted action chunk is supervised with an autoregressive
cross-entropy loss:
\begin{equation}
    \mathcal{L}_{\mathrm{act}}
    =
    -\frac{1}{K}
    \sum_{k=0}^{K-1}
    \log
    p_{\theta}
    \left(
        a_{t+k}^{*}
        \mid
        a_{t:t+k-1}^{*},
        I,\mathcal{M}_t,V_t,P_t
    \right).
    \label{eq:action_loss}
\end{equation}

\paragraph{Pose Loss.}
Let $(\mathbf t^\Delta,\mathbf q^\Delta)$ encode the ground-truth
inverse local transform
$\Delta E_{t\rightarrow t+K}^{\mathrm{w2c},*}$.
GT rotations are deterministically converted to canonical WXYZ
quaternions with $\|\mathbf q^\Delta\|_2=1$ and $q_w^\Delta\geq0$.
Following the component-wise regression of
Cambrian-P~\citep{yang2026cambrian} and VGGT~\citep{wang2025vggt},
we optimize
\begin{equation}
    \mathcal{L}_{\mathrm{pose}}
    =
    \sum_{s=1}^{S}
    w_s
    \left(
        \|\hat{\mathbf t}^{(s)}-\mathbf t^\Delta\|_1
        +
        \|\hat{\mathbf q}^{(s)}-\mathbf q^\Delta\|_1
    \right),
    \label{eq:pose_loss}
\end{equation}
where $w_s$ assigns larger weights to later refinement stages.
The quaternion term regresses the canonical target from the raw output;
it is not a geodesic or sign-invariant loss.
Predicted quaternions are normalized only during decoding,
not when computing the regression loss.

\paragraph{Directional Progress Loss.}
To discourage insufficient predicted translation along the target
direction, we complement the pose regression loss with a directional
progress constraint.
Specifically, we constrain the predicted translation's projection
onto the target direction in the inverse-extrinsic encoding.
Let $d=\|\mathbf t^\Delta\|_2$; for $d>\tau$, we define
\begin{equation}
    \mathcal{L}_{\mathrm{prog}}
    =
    \max\!\left(
        0,\;
        \rho d
        -
        \hat{\mathbf t}^{\top}
        \frac{\mathbf t^\Delta}{d}
    \right),
    \qquad d>\tau,
    \label{eq:progress_loss}
\end{equation}
where $\hat{\mathbf t}$ is the final-stage translation,
$\rho$ specifies the required projection ratio, and
$\tau$ filters negligible translations.
The loss is zero for $d\leq\tau$.
Both translation terms represent inverse-extrinsic components,
rather than camera-center displacements.
The complete formulation and hyperparameters appear in
Appendix~\ref{app:progress_loss}.

The overall objective is
\begin{equation}
    \mathcal{L}
    =
    \mathcal{L}_{\mathrm{act}}
    +
    \lambda_{\mathrm{pose}}\mathcal{L}_{\mathrm{pose}}
    +
    \lambda_{\mathrm{prog}}\mathcal{L}_{\mathrm{prog}}.
    \label{eq:total_loss}
\end{equation}
Here, $\lambda_{\mathrm{pose}}$ and $\lambda_{\mathrm{prog}}$ weight
pose regression and directional progress supervision, respectively.

\paragraph{Training and Inference.}
Initialized from StreamVLN~\citep{wei2025streamvln}, PoseVLN undergoes
supervised cold-start training followed by iterative closed-loop DAgger
refinement~\citep{wei2025streamvln,hao2026latentpilot}, without introducing
additional datasets.
Following LatentPilot's~\citep{hao2026latentpilot} training pipeline, PoseVLN further incorporates
pose-aware inputs and geometric supervision.
During rollout collection, predicted pose states are recursively propagated,
allowing pose errors to affect subsequent decisions and visited states.
Simulator camera poses are recorded for pose context and local
$\Delta$Pose supervision, and are not used as regular step-wise policy inputs
except for re-anchoring during DAgger expert recovery.
At inference, only the first observation defines the episode-relative origin
($P_0=I$); all subsequent poses are recursively predicted without ground-truth
poses or external odometry.
Further details on pose propagation and implementation are provided in
Appendices~\ref{app:pose_propagation} and~\ref{app:implementation},
respectively.

\section{Experiments}
\label{sec:experiments}
\begin{table*}[t]
\centering
\caption{
Comparison with state-of-the-art methods on the R2R-CE Val-Unseen split.
Pano., Odo., Depth, and S.RGB denote panoramic view, odometry,
depth, and single RGB observations, respectively.
All baseline results are reported from their respective papers.
}
\label{tab:r2r_main}

\setlength{\tabcolsep}{4.2pt}
\renewcommand{\arraystretch}{1.03}

\resizebox{\textwidth}{!}{
\begin{tabular}{l|cccc|c|cccc}
\toprule
\multirow{2}{*}{\textbf{Method}}
& \multicolumn{4}{c|}{\textbf{Observation}}
& \multirow{2}{*}{\textbf{DAgger}}
& \multicolumn{4}{c}{\textbf{R2R-CE Val-Unseen}} \\
\cmidrule(lr){2-5}
\cmidrule(lr){7-10}

& \textbf{Pano.}
& \textbf{Odo.}
& \textbf{Depth}
& \textbf{S.RGB}
&
& \textbf{NE}$\downarrow$
& \textbf{OS}$\uparrow$
& \textbf{SR}$\uparrow$
& \textbf{SPL}$\uparrow$ \\
\midrule

% ============================================================
% Panoramic + Odometry + Depth
% Sorted by SR
% ============================================================

HPN+DN {\scriptsize [ICCV21]}~\citep{krantz2021waypoint}
& \checkmark & \checkmark & \checkmark &
&
& 6.31 & 40.0 & 36.0 & 34.0 \\

CMA {\scriptsize [CVPR22]}~\citep{hong2022bridging}
& \checkmark & \checkmark & \checkmark &
&
& 6.20 & 52.0 & 41.0 & 36.0 \\

Sim2Sim {\scriptsize [ECCV22]}~\citep{krantz2022sim}
& \checkmark & \checkmark & \checkmark &
&
& 6.07 & 52.0 & 43.0 & 36.0 \\

VLN$\circlearrowright$BERT {\scriptsize [CVPR22]}~\citep{hong2022bridging}
& \checkmark & \checkmark & \checkmark &
&
& 5.74 & 53.0 & 44.0 & 39.0 \\

Ego$^2$-Map {\scriptsize [ICCV23]}~\citep{hong2023learning}
& \checkmark & \checkmark & \checkmark &
&
& 5.54 & 56.0 & 47.0 & 41.0 \\

DreamWalker {\scriptsize [ICCV23]}~\citep{wang2023dreamwalker}
& \checkmark & \checkmark & \checkmark &
&
& 5.53 & 59.0 & 49.0 & 44.0 \\

GridMM {\scriptsize [ICCV23]}~\citep{wang2023gridmm}
& \checkmark & \checkmark & \checkmark &
&
& 5.11 & 61.0 & 49.0 & 41.0 \\

Reborn {\scriptsize [CVPR22]}~\citep{an20221st}
& \checkmark & \checkmark & \checkmark &
&
& 5.40 & 57.0 & 50.0 & 46.0 \\

ScaleVLN {\scriptsize [ICCV23]}~\citep{wang2023scalevln}
& \checkmark & \checkmark & \checkmark &
&
& 4.80 & -- & 55.0 & 51.0 \\

ETPNav {\scriptsize [TPAMI24]}~\citep{an2023etpnav}
& \checkmark & \checkmark & \checkmark &
& \checkmark
& 4.71 & 65.0 & 57.0 & 49.0 \\

\midrule

% ============================================================
% Partial / Mixed Sensors
% Sorted by SR
% ============================================================

Seq2Seq {\scriptsize [ECCV20]}~\citep{krantz2020beyond}
& & & \checkmark & \checkmark
& \checkmark
& 7.77 & 37.0 & 25.0 & 22.0 \\

R2R-CMTP {\scriptsize [CVPR21]}~\citep{chen2021cmtp}
& \checkmark & \checkmark & \checkmark &
&
& 7.90 & 38.0 & 26.4 & 22.7 \\

InstructNav {\scriptsize [CoRL24]}~\citep{long2024instructnav}
& \checkmark & \checkmark & \checkmark &
&
& 6.89 & -- & 31.0 & 24.0 \\

CM2 {\scriptsize [CVPR22]}~\citep{georgakis2022cm2}
& & \checkmark & \checkmark & \checkmark
&
& 7.02 & 41.5 & 34.3 & 27.6 \\

LAW {\scriptsize [EMNLP21]}~\citep{raychaudhuri2021language}
& & \checkmark & \checkmark & \checkmark
& \checkmark
& 6.83 & 44.0 & 35.0 & 31.0 \\

WS-MGMap {\scriptsize [NIPS22]}~\citep{chen2022wsmgmap}
& & \checkmark & \checkmark & \checkmark
&
& 6.28 & 47.6 & 38.9 & 34.3 \\

MapNav {\scriptsize [ACL25]}~\citep{zhang2025mapnav}
& & \checkmark & \checkmark & \checkmark
& \checkmark
& 4.93 & 53.0 & 39.7 & 37.2 \\

NaVid-4D {\scriptsize [ICRA25]}~\citep{liu2025vid}
& & & \checkmark & \checkmark
&
& 5.99 & 55.7 & 43.8 & 37.1 \\

Sim2Real {\scriptsize [CoRL24]}~\citep{wang2024sim}
& & \checkmark & \checkmark& \checkmark
&
& 5.95 & 55.8 & 44.9 & 30.4 \\

COSMO {\scriptsize [ICCV25]}~\citep{zhang2025cosmo}
& \checkmark & & &
&
& -- & 56.0 & 47.0 & 40.0 \\

AO-Planner {\scriptsize [AAAI25]}~\citep{chen2025affordances}
& \checkmark & & \checkmark &
&
& 5.55 & 59.0 & 47.0 & 33.0 \\

g3D-LF {\scriptsize [CVPR25]}~\citep{wang2025g3d}
& & \checkmark & \checkmark & \checkmark
&
& 5.70 & 59.5 & 47.2 & 34.6 \\

NavMorph {\scriptsize [ICCV25]}~\citep{yao2025navmorph}
& & & \checkmark & \checkmark
&
& 5.75 & 56.9 & 47.9 & 33.2 \\

GA-VLN {\scriptsize [CVPR26]}~\citep{yang2026ga}
& & \checkmark & \checkmark & \checkmark
&
& 4.80 & 67.6 & 61.0 & 55.2 \\

\midrule

% ============================================================
% Single RGB
% Sorted by SR
% ============================================================

NaVid {\scriptsize [RSS24]}~\citep{zhang2024navid}
& & & & \checkmark
& \checkmark
& 5.47 & 49.1 & 37.4 & 35.9 \\

Uni-NaVid {\scriptsize [RSS25]}~\citep{zhang2024uni}
& & & & \checkmark
& \checkmark
& 5.58 & 53.3 & 47.0 & 42.7 \\

NaVILA {\scriptsize [RSS25]}~\citep{cheng2024navila}
& & & & \checkmark
&
& 5.22 & 62.5 & 54.0 & 49.0 \\

StreamVLN {\scriptsize [ICRA26]}~\citep{wei2025streamvln}
& & & & \checkmark
& \checkmark
& 4.98 & 64.2 & 56.9 & 51.9 \\

IDEAL-VLN {\scriptsize [ICML26]}~\citep{xininstruction}
& & & & \checkmark
&
& 4.62 & 64.9 & 59.8 & 52.9 \\

JanusVLN {\scriptsize [ICLR26]}~\citep{zeng2026janusvln}
& & & & \checkmark
& \checkmark
& 4.78 & 65.2 & 60.5 & 56.8 \\

LatentPilot {\scriptsize [ECCV26]}~\citep{hao2026latentpilot}
& & & & \checkmark
& \checkmark
& 4.41 & 66.3 & 62.0 & 58.0 \\

\midrule

\textbf{PoseVLN (Ours)}
& & & & \checkmark
& \checkmark
& \textbf{4.01}
& \textbf{69.4}
& \textbf{64.6}
& \textbf{59.4} \\

\bottomrule
\end{tabular}
}
\end{table*}

\subsection{Experimental Setup}

\paragraph{Simulation Environments and Metrics.}
Following recent continuous Vision-Language Navigation methods~\citep{
krantz2020beyond,
wei2025streamvln,
cheng2024navila,
zeng2026janusvln},
we evaluate PoseVLN on two standard VLN-CE benchmarks,
R2R-CE~\citep{anderson2018vln} and
RxR-CE English Val-Unseen~\citep{ku2020rxr}.
Both benchmarks are built upon Matterport3D environments and evaluated in
the Habitat simulator~\citep{savva2019habitat}.
We report results on the Val-Unseen splits.
For navigation evaluation, we report Navigation Error (NE), Oracle Success
Rate (OS), Success Rate (SR), Success weighted by Path Length (SPL), and
normalized Dynamic Time Warping (nDTW).
For pose evaluation, we report translational and rotational Relative Pose
Error (RPE) over different temporal horizons. All baseline results are taken
from their respective papers, with observation settings indicated in the
comparison tables.

\paragraph{Implementation Details.}
PoseVLN is initialized from StreamVLN~\citep{wei2025streamvln}, whose
navigation backbone is built on LLaVA-Video 7B with Qwen2-7B as the
language model.
We follow the same single-RGB streaming setting and train on trajectories
from R2R~\citep{anderson2018vln}, RxR~\citep{ku2020rxr}, and
ScaleVLN~\citep{wang2023scalevln}, together with closed-loop DAgger
rollouts~\citep{wei2025streamvln,hao2026latentpilot}, without introducing
additional datasets.
PoseVLN augments the backbone with pose-aware inputs and geometric
supervision.
Detailed implementation settings and training hyperparameters are provided
in Appendix~\ref{app:implementation}.

\begin{table*}[t]
\centering
\caption{
Comparison with state-of-the-art methods on the RxR-CE Val-Unseen split.
}
\label{tab:rxr_main}

\setlength{\tabcolsep}{4.0pt}
\renewcommand{\arraystretch}{1.03}

\resizebox{\textwidth}{!}{
\begin{tabular}{l|cccc|c|cccc}
\toprule
\multirow{2}{*}{\textbf{Method}}
& \multicolumn{4}{c|}{\textbf{Observation}}
& \multirow{2}{*}{\textbf{DAgger}}
& \multicolumn{4}{c}{\textbf{RxR-CE Val-Unseen}} \\
\cmidrule(lr){2-5}
\cmidrule(lr){7-10}

& \textbf{Pano.}
& \textbf{Odo.}
& \textbf{Depth}
& \textbf{S.RGB}
&
& \textbf{NE}$\downarrow$
& \textbf{SR}$\uparrow$
& \textbf{SPL}$\uparrow$
& \textbf{nDTW}$\uparrow$ \\
\midrule

% ============================================================
% Multi-sensor methods
% Sorted by SR
% ============================================================

CMA {\scriptsize [CVPR22]}~\citep{hong2022bridging}
& \checkmark & \checkmark & \checkmark &
&
& 8.76 & 26.5 & 22.1 & 47.0 \\

VLN$\circlearrowright$BERT {\scriptsize [CVPR22]}~\citep{hong2022bridging}
& \checkmark & \checkmark & \checkmark &
&
& 8.98 & 27.0 & 22.6 & 46.7 \\

Reborn {\scriptsize [CVPR22]}~\citep{an20221st}
& \checkmark & \checkmark & \checkmark &
&
& 5.98 & 48.6 & 42.0 & 63.3 \\

ETPNav {\scriptsize [TPAMI24]}~\citep{an2023etpnav}
& \checkmark & \checkmark & \checkmark &
& \checkmark
& 5.64 & 54.7 & 44.8 & 61.9 \\

\midrule

% ============================================================
% Partial / mixed sensor methods
% Sorted by SR
% ============================================================

LAW {\scriptsize [EMNLP21]}~\citep{raychaudhuri2021language}
& & \checkmark & \checkmark & \checkmark
& \checkmark
& 10.90 & 8.0 & 8.0 & 38.0 \\

Seq2Seq {\scriptsize [ECCV20]}~\citep{krantz2020beyond}
& & & \checkmark & \checkmark
& \checkmark
& 12.10 & 13.9 & 11.9 & 30.8 \\

Sim2Real {\scriptsize [CoRL24]}~\citep{wang2024sim}
& & \checkmark & \checkmark & \checkmark
&
& 8.79 & 25.5 & 18.1 & -- \\

NavMorph {\scriptsize [ICCV25]}~\citep{yao2025navmorph}
& & & \checkmark & \checkmark
&
& 8.85 & 30.8 & 22.8 & 44.2 \\

AO-Planner {\scriptsize [AAAI25]}~\citep{chen2025affordances}
& \checkmark & & \checkmark &
&
& 7.06 & 43.3 & 30.5 & 50.1 \\

GA-VLN {\scriptsize [CVPR26]}~\citep{yang2026ga}
& & \checkmark & \checkmark & \checkmark
&
& 5.88 & 55.4 & 45.2 & -- \\

\midrule

% ============================================================
% RGB-only methods
% Sorted by SR
% ============================================================

Uni-NaVid {\scriptsize [RSS25]}~\citep{zhang2024uni}
& & & & \checkmark
& \checkmark
& 6.24 & 48.7 & 40.9 & -- \\

NaVILA {\scriptsize [RSS25]}~\citep{cheng2024navila}
& & & & \checkmark
&
& 6.77 & 49.3 & 44.0 & 58.8 \\

StreamVLN {\scriptsize [ICRA26]}~\citep{wei2025streamvln}
& & & & \checkmark
& \checkmark
& 6.22 & 52.9 & 46.0 & 61.9 \\

JanusVLN {\scriptsize [ICLR26]}~\citep{zeng2026janusvln}
& & & & \checkmark
& \checkmark
& 6.06 & 56.2 & 47.5 & 62.1 \\

LatentPilot {\scriptsize [ECCV26]}~\citep{hao2026latentpilot}
& & & & \checkmark
& \checkmark
& 5.19 & 58.2 & 49.9 & 67.5 \\

IDEAL-VLN {\scriptsize [ICML26]}~\citep{xininstruction}
& & & & \checkmark
&
& 5.23 & 59.1 & 50.7 & 64.5 \\

\midrule

% ============================================================
% Ours
% ============================================================

\textbf{PoseVLN (Ours)}
& & & & \checkmark
& \checkmark
& \textbf{4.76} & \textbf{62.0} & \textbf{53.5} & \textbf{70.5} \\

\bottomrule
\end{tabular}
}
\end{table*}

\subsection{Main Results}

\paragraph{Results on R2R-CE.}
As shown in Table~\ref{tab:r2r_main}, we compare PoseVLN with a broad range of VLN methods on the R2R-CE Val-Unseen split under different observation settings.
PoseVLN operates solely from a single RGB stream, without panoramic observations, depth, or step-wise odometry.

PoseVLN achieves an NE of {4.01}, an OS of {69.4}, an SR of {64.6}, and an SPL of {59.4}, obtaining the best performance among all compared methods across all four metrics.
Compared with StreamVLN~\citep{wei2025streamvln}, which serves as our navigation backbone, PoseVLN improves SR from 56.9 to 64.6 and SPL from 51.9 to 59.4, corresponding to absolute gains of {7.7} and {7.5} points, while reducing NE from 4.98 to 4.01 and increasing OS from 64.2 to 69.4.
These gains indicate that explicitly modeling and propagating geometric state provides complementary spatial information beyond the visual-memory representation of the navigation backbone, improving both goal reaching and path efficiency without introducing additional sensing modalities.
Notably, these gains are achieved without using ground-truth pose or odometry during inference, supporting our hypothesis that camera pose can be learned and propagated as an internal geometric state for navigation.

\paragraph{Results on RxR-CE.}
As shown in Table~\ref{tab:rxr_main}, PoseVLN achieves
62.0\% SR, 53.5\% SPL, and 70.5 nDTW, with an NE of 4.76\,m
on RxR-CE Val-Unseen.
Compared with StreamVLN, PoseVLN improves SR, SPL, and nDTW
by 9.1, 7.5, and 8.6 points, respectively.
These improvements indicate higher navigation success,
better path efficiency, and closer agreement with the reference
trajectories on this challenging benchmark.
Nevertheless, the remaining unsuccessful episodes motivate
further analysis of accumulated pose errors and stopping behavior
over long navigation horizons.

\subsection{Ablation and Analysis}
\label{sec:ablation}

We conduct a series of ablation and diagnostic experiments to better
understand how pose-aware learning affects navigation performance and
geometric state estimation.

\paragraph{{Effect of Training Objectives and Pose Inputs.}}
We conduct cumulative ablations during cold-start training on R2R,
RxR, and ScaleVLN.
As shown in Table~\ref{tab:loss_ablation}, action-only replay reduces
the reproduced StreamVLN baseline from
{56.2\% to 53.5\% SR}.
This is consistent with replaying only the cold-start data shifting
the policy away from its later optimized state.
Adding {pose inputs and}
$\mathcal{L}_{\mathrm{pose}}$ recovers SR to
{56.3\%}, while introducing
$\mathcal{L}_{\mathrm{prog}}$ further
improves it to {57.2\%}.
Roughly one quarter of the training segments contain pure rotations
with little translation, which may partly explain the tendency toward
near-zero translation predictions under pose regression alone.
The progress loss discourages insufficient translation along the
target direction within the adopted pose parameterization.
{Additional diagnostics in
Appendix~\ref{app:progress_loss_analysis}
show that the final checkpoint trained with the progress loss has
a relative translation magnitude deficit of 8.65\%, compared with
79.23\% without it, while the fraction of predictions below 0.1\,m
decreases from 64.32\% to 4.61\%.}

We further apply DAgger refinement using the same aggregated
closed-loop rollout data.
The action-only baseline reaches
{57.0\% SR}, whereas full PoseVLN
achieves {64.6\%},
indicating that additional closed-loop training alone
cannot explain the final gain.
{
To separate auxiliary geometric learning from pose-conditioned navigation,
we train a variant with action loss, auxiliary $\Delta$Pose regression
($\mathcal{L}_{\mathrm{aux}}$), and directional progress loss
($\mathcal{L}_{\mathrm{prog}}$), but without current or historical
pose inputs during training or inference.
Here, $\mathcal{L}_{\mathrm{aux}}$ uses the same regression formulation
as $\mathcal{L}_{\mathrm{pose}}$, and both geometric losses retain
the weights and schedules used in full PoseVLN.
This variant achieves 57.5\% SR, improving over action-only training
by 0.5 percentage points but remaining 7.1 points below full PoseVLN.
Together, these results support a jointly trained geometric-state
framework rather than auxiliary pose regression alone:
geometric supervision helps the model interpret pose information,
pose-conditioned training enables effective use of that information,
and learned recursive feedback provides additional navigation gains.
}

{
We also evaluate Action-based Dead Reckoning (A-DR), which estimates
poses by composing nominal rigid-body transforms associated with
executed actions without measuring their actual displacement.
We provide A-DR poses as text prompts at inference to both the
action-only model and the model trained with action and auxiliary
geometric supervision.
SR decreases by 9.9 and 5.1 points relative to their respective
no-pose settings, suggesting that auxiliary geometric supervision
reduces sensitivity to these additional prompts, although direct
prompting still degrades performance.
For full PoseVLN, replacing predicted poses with A-DR estimates
through the existing pose-token interface yields 63.2\% SR,
compared with 64.6\% using learned pose feedback.
Thus, the full model can use action-integrated states, while learned
states provide better navigation performance in this evaluation.
}

\begin{table*}[t]
\centering
\captionsetup{font=footnotesize,labelfont=bf,skip=3pt}

% ================= Left Table =================
\begin{minipage}[t]{0.49\textwidth}
\vspace{0pt}
\centering
\caption{Geometric supervision and training ablations.}
\label{tab:loss_ablation}

\begingroup
\fontsize{6.3}{7.3}\selectfont
\setlength{\tabcolsep}{1.4pt}
\renewcommand{\arraystretch}{1.03}

\begin{tabular}{@{}lllrrr@{}}
\toprule
\textbf{Training}
& \textbf{Objective}
& \textbf{Pose}
& \textbf{SR (\%)}
& \textbf{SPL (\%)}
& \textbf{NE (m)} \\
\midrule

\textbf{StreamVLN}
& --
& None
& 56.2 & 50.0 & 5.10 \\

\midrule

\multirow{3}{*}{\textbf{w/o DAgger}}
& $\mathcal{L}_{\mathrm{act}}$
& None
& 53.5 & 48.0 & 5.25 \\

& $\mathcal{L}_{\mathrm{act}}+\mathcal{L}_{\mathrm{pose}}$
& Pred.
& 56.3 & 51.4 & 4.88 \\

& $\mathcal{L}_{\mathrm{act}}+\mathcal{L}_{\mathrm{pose}}+
\mathcal{L}_{\mathrm{prog}}$
& Pred.
& 57.2 & 52.2 & 4.71 \\

\midrule

\multirow{6}{*}{\textbf{+ DAgger}}
& $\mathcal{L}_{\mathrm{act}}$
& None
& 57.0 & 53.4 & 4.86 \\

& $\mathcal{L}_{\mathrm{act}}$
& A-DR
& 47.1 & 42.8 & 5.83 \\

& $\mathcal{L}_{\mathrm{act}}+\mathcal{L}_{\mathrm{aux}}+
\mathcal{L}_{\mathrm{prog}}$
& None
& 57.5 & 54.2 & 4.92 \\

& $\mathcal{L}_{\mathrm{act}}+\mathcal{L}_{\mathrm{aux}}+
\mathcal{L}_{\mathrm{prog}}$
& A-DR
& 52.4 & 48.9 & 5.45 \\

& $\mathcal{L}_{\mathrm{act}}+\mathcal{L}_{\mathrm{pose}}+
\mathcal{L}_{\mathrm{prog}}$
& Pred.
& \textbf{64.6} & \textbf{59.4}
& \textbf{4.01} \\

& $\mathcal{L}_{\mathrm{act}}+\mathcal{L}_{\mathrm{pose}}+
\mathcal{L}_{\mathrm{prog}}$
& A-DR
& 63.2 & 58.8 & 4.16 \\

\bottomrule
\end{tabular}
\endgroup
\end{minipage}%
\hfill
% ================= Right Table =================
\begin{minipage}[t]{0.49\textwidth}
\vspace{0pt}
\centering
\caption{Pose prediction comparison across horizons.
PoseVLN composes predictions, while pretrained VGGT
observes endpoint images.}
\label{tab:comparison_vggt}

\begingroup
\fontsize{7}{8.2}\selectfont
\setlength{\tabcolsep}{1.75pt}
\renewcommand{\arraystretch}{1.09}

\begin{tabular}{@{}crrrrr@{}}
\toprule
& \textbf{\#Pairs}
& \multicolumn{2}{c}{\textbf{Pos. RPE (m)}}
& \multicolumn{2}{c}{\textbf{Rot. RPE ($^\circ$)}} \\
\cmidrule(lr){3-4}
\cmidrule(lr){5-6}

\textbf{$H$}
& \textbf{Moving/All}
& \textbf{PoseVLN}
& \textbf{VGGT}
& \textbf{PoseVLN}
& \textbf{VGGT} \\
\midrule

4 & 20,836/27,912
& 0.071 & \textbf{0.015}
& 7.68 & \textbf{0.70} \\

8 & 21,277/26,069
& 0.145 & \textbf{0.065}
& \textbf{11.67} & 11.98 \\

12 & 20,253/24,228
& 0.242 & \textbf{0.215}
& \textbf{14.73} & 23.24 \\

16 & 18,660/22,387
& \textbf{0.362} & 0.453
& \textbf{17.40} & 31.35 \\

\bottomrule
\end{tabular}
\endgroup
\end{minipage}

\end{table*}

\begin{figure}[t]
    \centering
    \includegraphics[width=\linewidth]{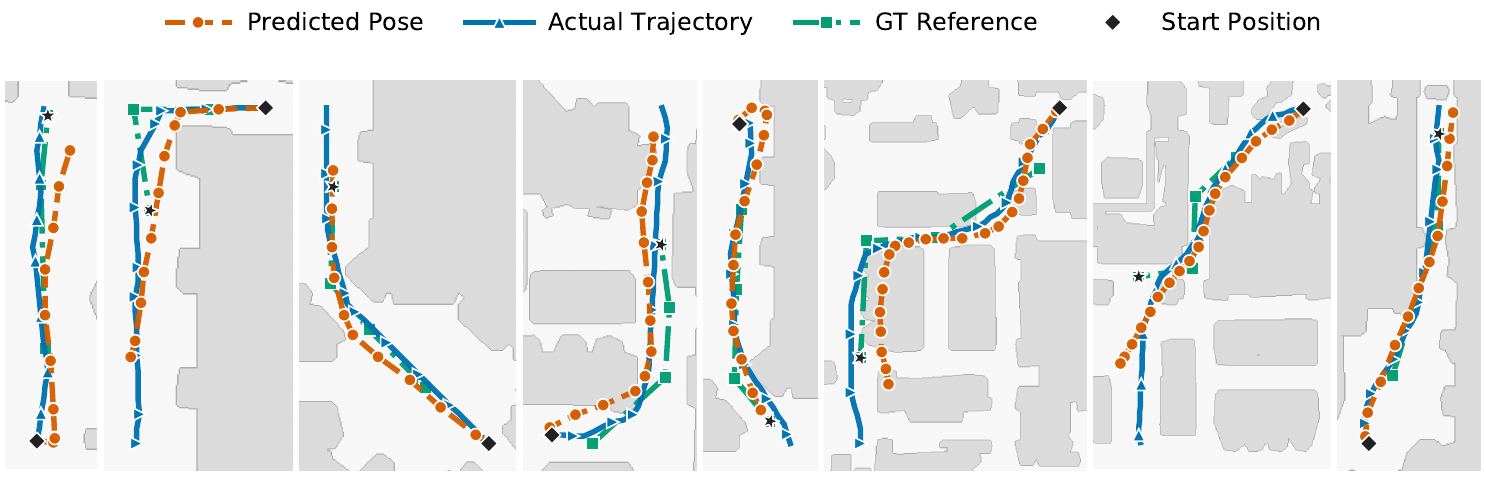}
    \caption{
    Qualitative comparison between predicted pose trajectories
    and actual navigation trajectories on R2R-CE Val-Unseen.
    }
    \label{fig:pose_trajectory}
    \vspace{-6pt}
\end{figure}

\paragraph{Pose prediction compared with VGGT.}
We compare PoseVLN with VGGT~\citep{wang2025vggt} over horizons
$H\in\{4,8,12,16\}$.
PoseVLN recursively composes successive $K=4$ predictions,
incorporating newly observed RGB frames, updated memory, and the
previously predicted pose at each prediction step.
VGGT instead performs direct two-view inference from the endpoint
image pair $[I_i,I_{i+H}]$.
Since VGGT's monocular two-view reconstruction is subject to global
scale ambiguity, translation errors are evaluated after scale alignment.
For each valid prediction--GT pair, both outputs are converted to local
$c2w$ displacements.
A non-negative translation scale is independently estimated as
$s^\ast=\max(0,\widehat{\mathbf t}^{\top}\mathbf t^{\mathrm{GT}}/
\|\widehat{\mathbf t}\|_2^2)$,
with the same per-pair alignment procedure applied to both methods.
This evaluation measures translation agreement up to scale rather
than metric displacement accuracy.
Position RPE is evaluated on pairs with GT displacement greater than
$0.05$\,m, while Rotation RPE is computed on all valid pairs.
All errors are micro-averaged over the corresponding valid pairs.
As shown in Table~\ref{tab:comparison_vggt}, VGGT achieves lower
Position RPE for $H=4,8,12$, whereas PoseVLN performs better at $H=16$.
For Rotation RPE, VGGT achieves lower error at $H=4$, while PoseVLN
is comparable or better at longer horizons.

\paragraph{Comparison between predicted and actual trajectories.}
Figure~\ref{fig:pose_trajectory} qualitatively compares the
recursively predicted pose trajectories with the actual navigation
trajectories on R2R-CE Val-Unseen.
The predicted trajectories generally capture the overall motion
direction, major turns, and trajectory shape.
In some long-horizon examples, the overall structure remains
recognizable despite noticeable endpoint drift, illustrating that
coarse trajectory consistency can coexist with accumulated
localization errors.
These examples suggest that PoseVLN captures meaningful geometric
structure over multiple navigation steps, while accurate
long-horizon pose propagation remains a challenge.

\paragraph{Pose accuracy in successful and failed navigation episodes.}
The relationship between geometric state estimation and navigation success
is examined by grouping R2R-CE Val-Unseen episodes according to their final
outcome ($\mathrm{SR}=1/0$).
Here, $H=4$ denotes the direct $K=4$ prediction, while $H=8,12,16$
are obtained by autoregressively composing successive $K=4$ pose transitions.
As shown on the left side of Table~\ref{tab:pose_analysis}, successful
episodes consistently exhibit lower pose errors across all horizons,
with larger absolute gaps under longer-horizon composition.
At $H=16$, the mean raw position error reaches $1.660$\,m for failed
episodes, nearly $3\times$ that of successful episodes ($0.554$\,m).
Similarly, translation-direction errors are lower in successful
episodes than in failed ones: $7.67^\circ$ versus $12.54^\circ$
at $H=4$, and $12.29^\circ$ versus $22.01^\circ$ at $H=16$.
These results indicate an association between navigation success
and more accurate geometric state estimation, particularly in
translation position and direction over long-horizon propagation.

\paragraph{Effect of pose inputs.}
We further study the contribution of memory and current pose information
through an inference-time input ablation.
All variants use the same trained PoseVLN checkpoint, with the corresponding
pose inputs masked only during evaluation.
As shown on the right of Table~\ref{tab:pose_analysis}, using both memory and
current poses achieves the best performance, with 64.6\% SR and 59.4\% SPL.
Removing either pose source consistently degrades navigation performance,
while removing both yields the lowest SR and SPL.
Masking the current pose leads to a slightly larger degradation than masking
memory poses, while the full model performs best across all reported metrics.
These results indicate that both current and memory pose states provide useful
geometric context under the jointly trained model.

\begin{table*}[t]
\centering

\caption{
Pose analysis on R2R-CE Val-Unseen.
Left: comparison of pose errors between successful and failed navigation episodes.
Right: ablation of memory and current pose inputs.
}
\label{tab:pose_analysis}

\vspace{2pt}

% ============================================================
% Left: Successful / Failed (unchanged)
% ============================================================
\begin{minipage}[t]{0.45\textwidth}
\centering
\vspace{0pt}

\scriptsize
\setlength{\tabcolsep}{3.2pt}
\renewcommand{\arraystretch}{1.2}

\resizebox{\linewidth}{!}{
\begin{tabular}{c c c c}
\toprule

\multicolumn{4}{c}{
\textit{Each entry: Successful / Failed (S/F)}
} \\

\midrule

\textbf{$H$} &
\textbf{Raw Pos. (m)} $\downarrow$ &
\textbf{Direction ($^\circ$)} $\downarrow$ &
\textbf{Rotation ($^\circ$)} $\downarrow$ \\

\midrule

4 &
\textbf{0.145} / 0.383 &
\textbf{7.67} / 12.54 &
\textbf{7.43} / 7.96 \\

8 &
\textbf{0.262} / 0.769 &
\textbf{8.78} / 17.72 &
\textbf{11.09} / 12.29 \\

12 &
\textbf{0.396} / 1.193 &
\textbf{10.67} / 19.29 &
\textbf{13.85} / 15.64 \\

16 &
\textbf{0.554} / 1.660 &
\textbf{12.29} / 22.01 &
\textbf{16.25} / 18.55 \\

\bottomrule
\end{tabular}
}
\end{minipage}
\hfill
% ============================================================
% Right: Pose Input Ablation
% ============================================================
\begin{minipage}[t]{0.54\textwidth}
\centering
\vspace{0pt}

\scriptsize
\setlength{\tabcolsep}{4.0pt}
\renewcommand{\arraystretch}{1.08}

\resizebox{\linewidth}{!}{
\begin{tabular}{l|cc|cccc}
\toprule

\multirow{2}{*}{\textbf{Method}}
&
\multicolumn{2}{c|}{\textbf{Pose Input}}
&
\multicolumn{4}{c}{\textbf{R2R-CE Val-Unseen}}
\\

\cmidrule(lr){2-3}
\cmidrule(lr){4-7}

&
\textbf{Hist.}
&
\textbf{Curr.}
&
\textbf{NE}$\downarrow$
&
\textbf{OS}$\uparrow$
&
\textbf{SR}$\uparrow$
&
\textbf{SPL}$\uparrow$
\\

\midrule

\textbf{PoseVLN}
& \checkmark
& \checkmark
& \textbf{4.01}
& \textbf{69.4}
& \textbf{64.6}
& \textbf{59.4}
\\

{w/ Action-DR Pose}
& {\checkmark}
& {\checkmark}
& {4.16}
& {65.7}
& {63.2}
& {58.8}
\\

w/o Memory Pose
&
& \checkmark
& 4.05
& 68.1
& 63.3
& 58.1
\\

w/o Current Pose
& \checkmark
&
& 4.15
& 68.1
& 62.9
& 57.9
\\

w/o Mem./Curr. Pose
&
&
& 4.11
& 67.7
& 62.9
& 57.7
\\

\bottomrule
\end{tabular}
}
\end{minipage}
\end{table*}

\paragraph{{Episode-Level Pose Error and Navigation Success.}}
{
To examine long-term geometric consistency beyond local pose
prediction, we evaluate accumulated errors over complete
R2R-CE Val-Unseen trajectories. Episodes are independently
ranked by translational ATE and rotational RMS error and
divided into five groups, with 368 episodes in each of
Q1--Q4 and 367 in Q5, covering all 1,839 episodes.
As shown in Table~\ref{tab:episode_pose_error},
SR decreases from 71.2\% to 45.0\% as mean ATE increases
from 0.330 to 8.126\,m. Similarly, SR decreases from
73.1\% to 46.0\% as rotational RMS error increases
from $8.00^\circ$ to $68.55^\circ$.
Although not strictly monotonic, these results associate
larger accumulated pose errors with lower navigation success,
supporting the value of maintaining useful geometric states
over extended trajectories.
}

\paragraph{{Learned Pose Feedback Under Execution Discrepancies.}}
{
We further compare learned $\Delta$Pose with Action-based Dead
Reckoning (Action-DR), which integrates nominal action transforms
through the same pose-token interface without retraining.
Table~\ref{tab:action_dr_collision} groups 1,839 matched tasks
by collision counts in Action-DR rollouts. While Action-DR
performs slightly better without collisions (69.5\% versus
68.2\% SR), learned pose feedback achieves higher SR in every
collision-containing group, reaching 23.5\% versus 12.3\%
when collisions exceed ten. In two representative scenes,
learned $\Delta$Pose also achieves lower unaligned ATE in
groups with at least three Action-DR collisions
(Appendix~\ref{app:episode_pose_analysis}).
These results support the navigation benefit of learned
geometric state feedback under execution discrepancies.
Nevertheless, PoseVLN does not explicitly correct accumulated
pose errors after execution, and long-horizon drift remains
a limitation.
}

\begin{table}[!htbp]
\centering
\captionsetup{
    font={footnotesize},
    labelfont={bf,sf},
    skip=3pt
}

\begin{minipage}[t]{0.48\linewidth}
\centering

\caption{Episode-level pose errors and navigation
success on R2R-CE Val-Unseen. }
\label{tab:episode_pose_error}

\footnotesize
\setlength{\tabcolsep}{3pt}
\renewcommand{\arraystretch}{1.05}
\begin{tabular}{@{}lrrrrr@{}}
\toprule
& & \multicolumn{2}{c}{Translation}
& \multicolumn{2}{c}{Rotation} \\
\cmidrule(lr){3-4}
\cmidrule(l){5-6}
\textbf{Group}
& \textbf{Episodes}
& \textbf{ATE}
& \textbf{SR}
& \textbf{RMS}
& \textbf{SR} \\
&
& \textbf{(m)}
& \textbf{(\%)}
& \textbf{($^\circ$)}
& \textbf{(\%)} \\
\midrule
Q1 & 368 & 0.330 & 71.2 & 8.00  & 73.1 \\
Q2 & 368 & 0.629 & 71.5 & 12.88 & 73.4 \\
Q3 & 368 & 1.008 & 68.5 & 19.45 & 66.6 \\
Q4 & 368 & 1.701 & 66.8 & 30.63 & 63.9 \\
Q5 & 367 & 8.126 & 45.0 & 68.55 & 46.0 \\
\midrule
All & 1,839 & -- & 64.6 & -- & 64.6 \\
\bottomrule
\end{tabular}
\end{minipage}
\hfill
\
\begin{minipage}[t]{0.50\linewidth}
\centering

\caption{Collision-stratified navigation success on
R2R-CE Val-Unseen. Collisions and median final horizontal
drift are measured from Action-DR rollouts.}
\label{tab:action_dr_collision}

\footnotesize
\setlength{\tabcolsep}{3pt}
\renewcommand{\arraystretch}{1.05}
\begin{tabular}{@{}lrrrr@{}}
\toprule
\textbf{Coll.}
& \textbf{Tasks}
& \shortstack{\textbf{A-DR}\\\textbf{SR (\%)}}
& \shortstack{\textbf{PoseVLN}\\\textbf{SR (\%)}}
& \shortstack{\textbf{A-DR Drift}\\\textbf{(m)}} \\
\midrule
0     & 946   & \textbf{69.5} & 68.2          & $\approx$0 \\
1--2  & 575   & 63.3          & \textbf{66.6} & 0.072 \\
3--5  & 169   & 60.4          & \textbf{63.3} & 0.275 \\
6--10 & 68    & 44.1          & \textbf{50.0} & 0.939 \\
$>10$ & 81    & 12.3          & \textbf{23.5} & 10.411 \\
\midrule
All   & 1,839 & 63.2          & \textbf{64.6} & -- \\
\bottomrule
\end{tabular}
\end{minipage}

\end{table}

\begin{figure}[t]
    \centering
    \includegraphics[width=\linewidth]{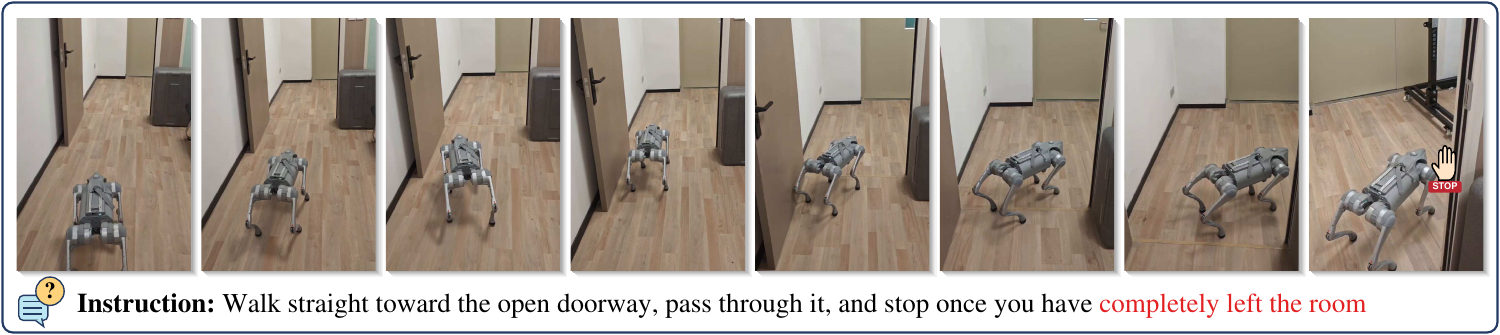}
    \caption{Real-world navigation of PoseVLN on a Unitree Go2.}
    \label{fig:real}
    \vspace{-6pt}
\end{figure}

\subsection{Real-World Deployment}

We further conduct a preliminary real-world deployment of PoseVLN on a
Unitree Go2 equipped with an onboard RGB camera, with inference performed
on a remote RTX 4090 server.
As shown in Figure~\ref{fig:real}, the robot successfully follows the
instruction, traverses the doorway, and stops after leaving the room.
{
We additionally conduct a quantitative evaluation across two real-world
scenes and six navigation task settings, with three trials per task.
PoseVLN successfully completes 11 of 18 trials (61.1\%), with
at least two successful trials in four of the six task settings.
Detailed results and trajectory visualizations are provided in
Appendix~\ref{app:real_world}.
}
This result demonstrates the applicability of PoseVLN to physical robot
navigation under real visual observations,
{
while broader evaluation across diverse environments and instructions
remains necessary.
}

\section{Limitations}
PoseVLN still has several limitations.
First, recursive pose composition can accumulate estimation errors and lead to
long-horizon drift.
{
Second, while the auxiliary-only ablation helps distinguish the contributions
of geometric supervision and recursively propagated pose inputs,
their interactions remain to be fully investigated.
}
Finally, the complementarity between learned $\Delta$Pose and
action-derived pose estimates remains to be further investigated.
Future work may explore more robust long-horizon pose propagation,
{
further analysis of the interaction between geometric supervision
and state representation,
}
and hybrid geometric estimation that combines learned pose transitions with
action-derived motion cues.
\section{Conclusion}
We introduce {PoseVLN}, a pose-aware VLN framework that connects past memory, current spatial state, and future motion through geometric state evolution.
PoseVLN organizes observations in an episode-relative coordinate frame and jointly predicts navigation actions and local $\Delta$Pose, whose recursive propagation maintains an evolving spatial state throughout navigation.
Using only single-RGB observations without step-wise ground-truth pose or odometry at inference, PoseVLN achieves strong performance on R2R-CE and RxR-CE.
These results demonstrate the effectiveness of camera pose as a compact geometric bridge for temporal state reasoning in VLN.
% \subsection*{AI use statement}

% In this work, we did not use generative AI tools for tasks requiring
% mandatory disclosure under the ICLR 2027 AI Policy for Authors.
% Tasks involving mathematical proofs and qualitative or thematic
% data analysis are not applicable to this work.
% Additionally, we used ChatGPT to improve the language and readability
% of the manuscript.
% We have reviewed all AI-assisted work. Specifically, the authors
% checked and revised the suggested edits to ensure that they preserved
% the intended technical meaning and accurately reflected the methods
% and experimental results.
% We take responsibility for the final content of this work, including
% text, claims, or artifacts produced with the aid of generative AI.

% \subsection*{Reproducibility statement}

% To facilitate reproducibility, we describe the architecture, pose
% representation, training objectives, and inference procedure of
% PoseVLN in Section~\ref{sec:method}.
% The datasets, implementation details, training settings, and
% evaluation protocols are provided in Section~\ref{sec:experiments} and Appendices.
% The ablation studies and pose-error analyses in
% Section~\ref{sec:experiments} further document the contributions
% of the main components and the evaluation of pose prediction.

% LeapWing uses numeric natbib citations and the original bibliography data.
\bibliographystyle{plainnat}
\bibliography{ref}

\clearpage
\appendix
\begin{center}
    {\Large\bfseries Appendix}
\end{center}
\vspace{0.5em}
This appendix provides additional details of PoseVLN, including the
pose representation and recursive geometric state propagation,
closed-loop DAgger collection, the Directional Progress Loss,
training sample construction, and implementation
details. We further clarify how geometric supervision is used during
training. Simulator-recorded camera poses are used to construct
pose-aware training context and local $\Delta$Pose targets, whereas
at inference only the first observation defines the episode-relative
origin and all subsequent pose states are recursively propagated from
model predictions without ground-truth pose or external odometry.

% ============================================================
\section{Pose Representation and Geometric State Propagation}
\label{app:pose_geometry}

% ------------------------------------------------------------
\subsection{Episode-relative Camera Coordinates}
\label{app:episode_coordinates}

Using column-vector coordinates, let the simulator camera-to-world
transformation of frame $t$ be
\begin{equation}
    T_t^{\mathrm{c2w}}
    =
    \begin{bmatrix}
        R_t & p_t\\
        0^\top & 1
    \end{bmatrix}
    \in SE(3),
\end{equation}
where $R_t$ and $p_t$ denote the camera orientation and center in the
simulator global coordinate system.

We use the first camera frame as the episode reference:
\begin{equation}
    \bar T_t^{\mathrm{c2w}}
    =
    \left(T_0^{\mathrm{c2w}}\right)^{-1}
    T_t^{\mathrm{c2w}}.
    \label{eq:app_episode_pose}
\end{equation}
By construction, $\bar T_0^{\mathrm{c2w}}=I_4$.
This rigid coordinate change preserves metric scale; no additional
trajectory-scale normalization is applied.
The main-text trajectory-level state is
\begin{equation}
    P_t
    \equiv
    \bar T_t^{\mathrm{c2w}},
    \label{eq:app_main_pose}
\end{equation}
which maps camera coordinates to the episode frame.

For pose tokenization, we use the inverse world-to-camera representation:
\begin{equation}
    E_t^{\mathrm{w2c}}
    =
    P_t^{-1}
    =
    \left(
        \bar T_t^{\mathrm{c2w}}
    \right)^{-1}.
    \label{eq:app_w2c_pose}
\end{equation}
Thus, the trajectory state is stored in c2w form, while its pose token
encodes the corresponding w2c rotation and translation.

% ------------------------------------------------------------
\subsection{Local \texorpdfstring{$\Delta$}{Delta}Pose Supervision}
\label{app:delta_pose}

For a current timestep $i$ and a future timestep $j=i+K$, the local
camera transformation is
\begin{equation}
    \Delta T_{i\rightarrow j}^{\mathrm{c2w}}
    =
    \left(
        \bar T_i^{\mathrm{c2w}}
    \right)^{-1}
    \bar T_j^{\mathrm{c2w}}.
    \label{eq:app_delta_c2w}
\end{equation}
Writing
\begin{equation}
    \bar T_i^{\mathrm{c2w}}
    =
    \begin{bmatrix}
        \bar R_i & \bar p_i\\
        0^\top & 1
    \end{bmatrix},
    \qquad
    \bar T_j^{\mathrm{c2w}}
    =
    \begin{bmatrix}
        \bar R_j & \bar p_j\\
        0^\top & 1
    \end{bmatrix},
\end{equation}
gives
\begin{equation}
    R_{ij}^{\mathrm{rel}}
    =
    \bar R_i^\top\bar R_j,
    \qquad
    p_{ij}^{\mathrm{rel}}
    =
    \bar R_i^\top
    (\bar p_j-\bar p_i).
    \label{eq:app_relative_components}
\end{equation}
Here, $p_{ij}^{\mathrm{rel}}$ is the displacement of the future camera
center expressed in the current camera frame.
The main-text transition is
$\Delta P_{i\rightarrow j}=\Delta T_{i\rightarrow j}^{\mathrm{c2w}}$.

Substituting Eq.~\ref{eq:app_episode_pose} into
Eq.~\ref{eq:app_delta_c2w} yields
\begin{equation}
    \Delta T_{i\rightarrow j}^{\mathrm{c2w}}
    =
    \left(
        T_i^{\mathrm{c2w}}
    \right)^{-1}
    T_j^{\mathrm{c2w}},
\end{equation}
showing that the local transformation is independent of the arbitrary
global origin and orientation.

We supervise the CameraHead using the inverse local transform:
\begin{equation}
    \Delta E_{i\rightarrow j}^{\mathrm{w2c}}
    =
    \left(
        \Delta T_{i\rightarrow j}^{\mathrm{c2w}}
    \right)^{-1}
    =
    E_j^{\mathrm{w2c}}
    \left(E_i^{\mathrm{w2c}}\right)^{-1}.
    \label{eq:app_delta_w2c}
\end{equation}
This transform maps current-camera coordinates to future-camera
coordinates.
Its rotation and translation satisfy
\begin{equation}
    R_{ij}^{\mathrm{w2c}}
    =
    (R_{ij}^{\mathrm{rel}})^\top,
    \qquad
    \mathbf t_{ij}^{\Delta}
    =
    -(R_{ij}^{\mathrm{rel}})^\top p_{ij}^{\mathrm{rel}}.
    \label{eq:app_inverse_components}
\end{equation}
Accordingly, the translation supervised by the pose and Directional
Progress objectives is an inverse-extrinsic component, rather than
the camera-center displacement $p_{ij}^{\mathrm{rel}}$.

% ------------------------------------------------------------
\subsection{9D Camera Parameterization}
\label{app:camera_parameterization}

We reuse the VGGT CameraHead architecture and follow the component-wise
pose regression of Cambrian-P and VGGT
\citep{yang2026cambrian,wang2025vggt}.
Our encoding explicitly uses WXYZ quaternion ordering:
\begin{equation}
    g_i
    =
    [\mathbf t_i,\mathbf q_i,\bar f^h,\bar f^w]
    =
    [t_x,t_y,t_z,q_w,q_x,q_y,q_z,\bar f^h,\bar f^w]
    \in\mathbb{R}^{9}.
    \label{eq:app_9d_camera}
\end{equation}
For trajectory-level tokens, $\mathbf t_i$ and $\mathbf q_i$ encode
the translation and rotation of $E_i^{\mathrm{w2c}}=P_i^{-1}$.
Local targets use the same encoding layout for
$\Delta E_{i\rightarrow j}^{\mathrm{w2c}}$.

\paragraph{Canonical Quaternion Regression.}
For a GT rotation $R^*$, we deterministically construct the target
\begin{equation}
    \mathbf q^*
    =
    \operatorname{Canon}
    \left(\operatorname{Quat}_{\mathrm{WXYZ}}(R^*)\right),
    \qquad
    \|\mathbf q^*\|_2=1,\quad q_w^*\geq0,
    \label{eq:app_canonical_quaternion}
\end{equation}
where canonicalization standardizes the quaternion sign.
The CameraHead outputs an unconstrained vector
$\hat{\mathbf q}\in\mathbb{R}^4$, and the rotation term directly
minimizes $\|\hat{\mathbf q}-\mathbf q^*\|_1$ at each refinement stage.
This is component-wise regression against a canonical target,
rather than a geodesic or sign-invariant rotation loss.
Predicted quaternions are normalized during geometric decoding,
not before computing this regression loss.

\paragraph{Fixed Field of View.}
The two FoV entries are retained for compatibility with the 9D
CameraHead interface.
They are overwritten with known sensor constants during training
and inference, contribute no effective FoV loss, and are excluded
from geometric state propagation.

% ------------------------------------------------------------
\subsection{Recursive Pose Propagation}
\label{app:pose_propagation}

At inference, PoseVLN initializes the episode coordinate system with
$\hat P_0=P_0=I_4$, without querying a GT camera pose.
For the final-stage quaternion output, geometric decoding applies
\begin{equation}
    \bar{\mathbf q}
    =
    \frac{\hat{\mathbf q}}
    {\max(\|\hat{\mathbf q}\|_2,\epsilon)},
    \qquad
    \hat R
    =
    \operatorname{QuatToMat}_{\mathrm{WXYZ}}(\bar{\mathbf q}).
    \label{eq:app_quaternion_decoding}
\end{equation}
The decoded rotation and predicted translation form
$\Delta\hat E_{i\rightarrow j}^{\mathrm{w2c}}$.
We invert this transform before state composition:
\begin{equation}
    \Delta\hat T_{i\rightarrow j}^{\mathrm{c2w}}
    =
    \left(
        \Delta\hat E_{i\rightarrow j}^{\mathrm{w2c}}
    \right)^{-1}.
    \label{eq:app_pred_delta_c2w}
\end{equation}
For decoded w2c components $(\hat R,\hat{\mathbf t})$, the inverse
has rotation $\hat R^\top$ and translation
$-\hat R^\top\hat{\mathbf t}$.
The trajectory-level state is updated by
\begin{equation}
    \hat P_j
    =
    \hat P_i
    \Delta\hat T_{i\rightarrow j}^{\mathrm{c2w}}.
    \label{eq:app_pose_update}
\end{equation}

Equivalently, under the w2c convention,
\begin{equation}
    \hat E_j^{\mathrm{w2c}}
    =
    \Delta\hat E_{i\rightarrow j}^{\mathrm{w2c}}
    \hat E_i^{\mathrm{w2c}}.
    \label{eq:app_w2c_update}
\end{equation}

The propagated state $\hat P_j$ is paired with the new observation $V_j$
and stored in memory:
\begin{equation}
    \mathcal{M}_t
    =
    \{
        (V_{h_1},\hat P_{h_1}),
        \ldots,
        (V_{h_{N_h}},\hat P_{h_{N_h}})
    \}.
\end{equation}
Its pose token is constructed from
$\hat E_j^{\mathrm{w2c}}=\hat P_j^{-1}$.
Thus, all non-initial current and memory states are model-predicted.

The resulting state evolution is
\begin{equation}
    \hat P_0
    \xrightarrow{\Delta\hat P_{0\rightarrow K}}
    \hat P_K
    \xrightarrow{\Delta\hat P_{K\rightarrow 2K}}
    \hat P_{2K}
    \rightarrow \cdots .
\end{equation}
Pose errors therefore accumulate under the same recursive propagation
used for navigation evaluation.

% ============================================================
\section{Closed-Loop DAgger and Pose Collection}
\label{app:dagger_collection}

We collect additional training trajectories through closed-loop policy
rollouts in the Habitat simulator.
During collection, the policy executes predicted action chunks while its
trajectory-level pose state is recursively propagated from predicted local
$\Delta$Pose according to Appendix~\ref{app:pose_propagation}.
Consequently, accumulated pose errors can affect subsequent navigation
decisions during data collection.

For each simulator state used to produce an RGB observation, we record the
corresponding ground-truth camera pose $T_t^{\mathrm{c2w}}$.
Local pose targets are subsequently constructed from these temporally aligned
camera poses following Appendix~\ref{app:delta_pose}.
Each executed action is logged together with the RGB observation and camera
pose at the simulator state from which that action is issued, ensuring
temporal alignment among observations, actions, and pose supervision.

\paragraph{Expert Intervention.}
During a rollout, we compute the minimum distance $d_t$ between the current
agent position and the reference path.
The Habitat shortest-path expert takes control when $d_t>1.0$\,m and returns
control to the policy when $d_t<0.5$\,m.
Expert takeover is also triggered when the normalized action-error score
exceeds $0.5$, the accumulated action errors exceed $8$, or the policy
produces an invalid \texttt{Stop} action.

\paragraph{Pose Re-anchoring.}
When expert correction is triggered, the internally propagated pose state is
re-anchored to the current simulator camera pose:
\begin{equation}
    \hat P_t
    \leftarrow
    \left(
        T_0^{\mathrm{c2w}}
    \right)^{-1}
    T_t^{\mathrm{c2w}}.
    \label{eq:app_dagger_reanchor}
\end{equation}
Before control is returned to the policy, unexecuted action predictions and
stale recurrent states are discarded, and closed-loop action and pose
prediction resumes from the corrected state.

This re-anchoring is used only during training-data collection and is never
applied during evaluation or inference.
A corrected rollout therefore contains a policy-generated segment, an expert
recovery segment, and a subsequent policy-generated segment.
The simulator camera trajectory is recorded independently throughout the
rollout and is later used to construct ground-truth pose context and
$\Delta$Pose supervision for training.

\section{Directional Progress Loss}
\label{app:progress_loss}

The basic pose objective supervises translation and rotation independently.
We additionally introduce a directional constraint to discourage local
translation predictions with insufficient progress along the ground-truth
translation direction.

For the $m$-th valid local-motion target, let
$\mathbf t_m^\Delta$ denote the ground-truth local w2c translation and
$d_m=\|\mathbf t_m^\Delta\|_2$ its magnitude.
For samples satisfying $d_m>\tau$, the normalized local translation direction
is
\begin{equation}
    \mathbf u_m^\Delta
    =
    \frac{\mathbf t_m^\Delta}{d_m}.
\end{equation}

Let $\hat{\mathbf t}_m$ denote the translation predicted by the final
CameraHead refinement.
Its signed progress along the ground-truth direction is
\begin{equation}
    s_m^{\mathrm{prog}}
    =
    \hat{\mathbf t}_m^\top
    \mathbf u_m^\Delta.
\end{equation}
The complete Directional Progress Loss is
\begin{equation}
    \mathcal{L}_{\mathrm{prog}}
    =
    \frac{
        \sum_m
        \mathbb{I}(d_m>\tau)
        \max
        \left(
            0,\,
            \rho d_m-s_m^{\mathrm{prog}}
        \right)
    }
    {
        \sum_m
        \mathbb{I}(d_m>\tau)
        +
        \epsilon
    }.
    \label{eq:app_progress_loss}
\end{equation}
We use $\tau=0.1\,\mathrm{m}$ and $\rho=0.8$.

This auxiliary objective penalizes predictions with insufficient progress
along the ground-truth local translation direction, including near-zero or
oppositely directed predictions.
Importantly, $\mathcal{L}_{\mathrm{prog}}$ is a local geometric constraint
on the pose branch and should not be interpreted as task-level progress
toward the navigation goal.

% ============================================================
\section{Training Sample Construction}
\label{app:data}

We construct training samples from trajectories in R2R, RxR, ScaleVLN, and
the additional closed-loop rollouts described in
Appendix~\ref{app:dagger_collection}. For each trajectory, we sample an
action window of length $N=16$ and predict an action chunk every $K=4$ steps,
resulting in $M=N/K=4$ prediction segments per complete window.

For a window starting at timestep $s$, the $m$-th current state is
\begin{equation}
i_m=s+mK,
\qquad
m\in\{0,\ldots,M-1\},
\end{equation}
with future target
\begin{equation}
j_m
=
\min(i_m+K,T_{\mathrm{end}}).
\end{equation}
Each valid segment provides
\begin{equation}
V_{i_m}
\longrightarrow
\left(
A_{i_m:j_m-1},
\Delta P_{i_m\rightarrow j_m}
\right).
\label{eq:app_training_segment}
\end{equation}
The local transition is encoded as its inverse world-to-camera transform for
CameraHead supervision.

The pose annotations associated with the current observation and historical
memory observations are expressed in the episode-relative coordinate system
and encoded into pose latents. For non-initial windows, we additionally
sample at most $N_h=8$ historical observations:
\begin{equation}
\mathcal{M}_{i_m}
=
\left\{
\left(
V_{h_n},P_{h_n}
\right)
\right\}_{n=1}^{N_h}.
\end{equation}
These historical observation-pose pairs provide navigation context for the
current prediction segment but do not receive independent action or
local-pose supervision.

The closed-loop rollouts expose the policy to observation and action
sequences induced by its own predictions. The synchronized camera trajectory
provides geometrically aligned pose annotations and local $\Delta$Pose
targets for constructing the corresponding training samples.

% \begin{wrapfigure}{r}{0.46\linewidth}
%     \vspace{-8pt}
%     \centering
%     \includegraphics[width=\linewidth]{image/flywheel_sr2.pdf}
%     \caption{
%     Effect of flywheel training.
%     }
%     \label{fig:flywheel}
%     \vspace{-10pt}
% \end{wrapfigure}

% \paragraph{Effect of flywheel iterations.}
% We study the effect of iterative flywheel training following~\citep{hao2026latentpilot}.
% Starting from the StreamVLN backbone (56.17\% SR), cold-start training with the
% pose module improves SR to 57.21\%.
% Successive collect--train--select iterations further increase performance,
% reaching 64.63\% at Round 5 before slightly dropping to 63.91\% at Round 6
% (Fig.~\ref{fig:flywheel}).
% Overall, the flywheel process substantially improves the cold-start model without introducing additional datasets beyond those used for initial training.

% ============================================================
\section{Implementation Details}
\label{app:implementation}

PoseVLN is built upon StreamVLN under the single-RGB streaming setting.
The navigation backbone is initialized from the corresponding StreamVLN
checkpoint, while the pose encoder and VGGT CameraHead are initialized from
scratch.

We use trajectory windows of length $N=16$, at most $N_h=8$ memory
observations, and an action-prediction horizon of $K=4$.
The pose encoder maps the 9D camera representation to the
3,584-dimensional language-model embedding space.
The VGGT CameraHead operates on a 2,048-dimensional representation with
four transformer blocks, 16 attention heads, and four iterative refinements.

Training is performed on four GPUs using AdamW with a learning rate of
$5\times10^{-6}$, zero weight decay, a cosine schedule with $3\%$ warmup,
BF16 precision, gradient checkpointing, and DeepSpeed ZeRO-2.
The per-device batch size is $1$ with $32$ gradient-accumulation steps,
giving an effective batch size of $128$.

The action and pose loss weights are both $1.0$.
Within the pose objective, translation and quaternion regression
each use unit weight.
The inherited FoV weight is $0.5$, but its loss is zero because
the predicted FoV entries are overwritten with the target constants.
For the Directional Progress Loss, we use
$\tau=0.1\,\mathrm{m}$ and $\rho=0.8$.
Its coefficient is linearly increased from $0.05$ to $0.5$ during the first
$1{,}920$ supervised-training steps and fixed to $0.2$ during subsequent
DAgger refinement.

Unless otherwise specified, the remaining settings follow
StreamVLN~\citep{wei2025streamvln}.

\noindent\textcolor{blue}{
Appendices F--N present additional experiments, analyses, and
clarifications included in the revised manuscript during the rebuttal period.
}

\section{Scene-Level Geometric Error Under Execution Discrepancies}
\label{app:episode_pose_analysis}

The main text reports episode-level pose errors and navigation
success (Table~\ref{tab:episode_pose_error}) and collision-stratified
navigation results (Table~\ref{tab:action_dr_collision}).
Here, we compare unaligned translational ATE of learned
$\Delta$Pose and Action-based Dead Reckoning (Action-DR)
in two representative scenes. Groups are defined by Action-DR
collision counts. The variants share task identities and may
follow different executed trajectories.

\begin{table}[t]
    \centering
    \footnotesize
    \setlength{\tabcolsep}{3pt}
    \renewcommand{\arraystretch}{1.05}

    \caption{
        Collision-stratified mean episode-level translational
        ATE (m, lower is better) in two representative scenes.
        Groups are defined by Action-DR collision counts.
        Both variants use matched tasks without scale alignment.
    }
    \label{tab:action_dr_scene_ate}

    \begin{tabular}{@{}llrrr@{}}
        \toprule
        \textbf{Scene}
        & \shortstack{\textbf{Action-DR}\\\textbf{collisions}}
        & \textbf{Episodes}
        & \shortstack{\textbf{Action-DR}\\\textbf{ATE (m)}}
        & \shortstack{\textbf{Learned $\Delta$Pose}\\\textbf{ATE (m)}} \\
        \midrule
        EU6Fwq7SyZv
        & 0       & 61  & \textbf{0.318} & 1.099 \\
        & 1--2    & 39  & \textbf{0.309} & 1.484 \\
        & $\geq$3 & 32  & 18.249 & \textbf{16.061} \\
        & \textbf{Overall} & \textbf{132}
        & \textbf{4.662} & 4.840 \\
        \midrule
        x8F5xyUWy9e
        & 0       & 53  & \textbf{0.263} & 0.989 \\
        & 1--2    & 18  & \textbf{0.153} & 1.710 \\
        & $\geq$3 & 31  & 10.464 & \textbf{6.377} \\
        & \textbf{Overall} & \textbf{102}
        & 3.344 & \textbf{2.754} \\
        \bottomrule
    \end{tabular}
\end{table}

Table~\ref{tab:action_dr_scene_ate} shows lower ATE for Action-DR
in collision-free and low-collision groups. For groups with at least
three Action-DR collisions, learned $\Delta$Pose achieves lower
ATE in both scenes (16.061 versus 18.249\,m and 6.377 versus
10.464\,m). The overall ATE comparison differs across scenes.
Together with the navigation results in
Table~\ref{tab:action_dr_collision}, these findings show the
value of learned geometric feedback under larger execution
discrepancies.

\section{Translation Underestimation and the Directional Progress Loss}
\label{app:progress_loss_analysis}

We analyze translation underestimation and the effects of
Directional Progress Loss on translation magnitudes and physical
camera displacement, complementing the navigation ablation in
Table~\ref{tab:loss_ablation}.

\paragraph{Motivation and Experimental Setup.}
During exploratory R2R-train and RxR-train training with
$\mathcal{L}_{\mathrm{act}}$ and $\mathcal{L}_{\mathrm{pose}}$,
predicted translation magnitudes decreased at later checkpoints,
which can contract recursively propagated trajectories.
We compare intermediate and final checkpoints without the
progress loss and a final checkpoint with the loss, using the
same three checkpoints for magnitude and displacement analyses.

For each evaluated local prediction--target pair, let
$\hat{d}_m = \|\hat{\mathbf{t}}_m\|_2$ and
$d_m = \|\mathbf{t}^{*}_m\|_2$ denote the predicted and GT
translation magnitudes, respectively.
We report their means, $\bar{d}_{\mathrm{pred}}$ and
$\bar{d}_{\mathrm{GT}}$, together with the difference of means
and relative difference:
\begin{equation}
    b = \bar{d}_{\mathrm{pred}}-\bar{d}_{\mathrm{GT}},
    \qquad
    b_{\mathrm{rel}}
    = \frac{\bar{d}_{\mathrm{pred}}-\bar{d}_{\mathrm{GT}}}
           {\bar{d}_{\mathrm{GT}}}\times 100\%.
\end{equation}
Negative values indicate aggregate magnitude underestimation.
We also report the percentage of predictions with
$\hat{d}_m < 0.1\,\mathrm{m}$.
Each row includes the GT mean for its evaluated samples.

\begin{table}[t]
    \centering
    \small
    \setlength{\tabcolsep}{4pt}
    \caption{
        Translation underestimation and physical camera displacement
        diagnostics using the same three model checkpoints.
        \textbf{Top:} Translation magnitude statistics from
        exploratory training on R2R-train and RxR-train.
        Differences are computed between mean magnitudes,
        not per-sample absolute errors.
        \textbf{Bottom:} Physical displacement errors on a
        supplementary subset of 10 episodes and 157 four-step
        segments. Directional errors use 132 valid segments.
        The supplementary evaluations use independently generated
        rollouts for each checkpoint.
    }
    \label{tab:progress_translation_analysis}

    \begin{tabular}{@{}lrrrrr@{}}
        \toprule
        \textbf{Model}
        & \shortstack{\textbf{Mean Pred.}\\\textbf{(m)}}
        & \shortstack{\textbf{Mean GT}\\\textbf{(m)}}
        & \shortstack{\textbf{Difference}\\\textbf{(m)}}
        & \shortstack{\textbf{Relative}\\\textbf{Difference (\%)}}
        & \shortstack{\textbf{Pred. $<0.1$ m}\\\textbf{(\%)}} \\
        \midrule
        \shortstack[l]{Without progress,\\intermediate}
        & 0.3410 & 0.6501 & $-0.3091$ & $-47.55$ & 18.40 \\
        \shortstack[l]{Without progress,\\final}
        & 0.1356 & 0.6527 & $-0.5171$ & $-79.23$ & 64.32 \\
        \shortstack[l]{\textbf{With progress,}\\\textbf{final}}
        & \textbf{0.5778}
        & 0.6325
        & $\mathbf{-0.0547}$
        & $\mathbf{-8.65}$
        & \textbf{4.61} \\
        \bottomrule
    \end{tabular}

    \vspace{7pt}

    \begin{tabular}{@{}lrrr@{}}
        \toprule
        \textbf{Checkpoint}
        & \shortstack{\textbf{Mean Error}\\\textbf{(m)}}
        & \shortstack{\textbf{RMSE}\\\textbf{(m)}}
        & \shortstack{\textbf{Direction Error}\\\textbf{($^\circ$)}} \\
        \midrule
        Without progress (1920)
        & 0.3523 & 0.4153 & 16.64 \\
        Without progress (2560)
        & 0.6125 & 0.6989 & 88.03 \\
        \textbf{With progress (1920)}
        & \textbf{0.1736}
        & \textbf{0.2009}
        & \textbf{11.01} \\
        \bottomrule
    \end{tabular}
\end{table}

\paragraph{Observed Translation Underestimation.}
Table~\ref{tab:progress_translation_analysis} shows that,
without the progress loss, mean predicted translation decreases
from 0.3410 to 0.1356\,m while GT means remain near 0.65\,m.
The magnitude deficit increases from 47.55\% to 79.23\%,
and predictions below 0.1\,m rise from 18.40\% to 64.32\%.

With the progress loss, the final checkpoint predicts 0.5778\,m
against a GT mean of 0.6325\,m, reducing the deficit to 8.65\%
and predictions below 0.1\,m to 4.61\%. Together with the
navigation ablation, this supports its effectiveness in mitigating
translation underestimation.

\paragraph{Physical Camera Displacement Analysis.}
To further evaluate geometric accuracy in physical space,
we examine the same checkpoints using camera-center displacement
rather than inverse-extrinsic translation parameters.
The predicted local world-to-camera transformation is converted
into relative camera-center motion:
\begin{equation}
    \hat{\mathbf{p}}
    = -\hat{\mathbf{R}}^{\top}\hat{\mathbf{t}},
    \qquad
    \mathbf{p}^{*}
    =
    \left[
    (T_i^{*})^{-1}T_{i+4}^{*}
    \right]_{:3,3},
\end{equation}
where $\hat{\mathbf{R}}$ and $\hat{\mathbf{t}}$ denote the
predicted local world-to-camera rotation and translation,
and $T_i^{*}$ is the GT camera-to-world pose.
This conversion incorporates the predicted rotation when
estimating relative physical camera displacement.

For each segment, we measure the displacement vector error
\begin{equation}
    e_{\mathrm{disp}}
    = \|\hat{\mathbf{p}}-\mathbf{p}^{*}\|_2.
\end{equation}
We report its mean and root-mean-square error (RMSE),
together with the angle between the predicted and GT
displacement vectors.

The supplementary evaluation covers 10 common episode identities
and 157 four-step segments for each checkpoint.
All 157 segments are included in the displacement error
statistics. Directional errors are computed over 132 common
segments with GT displacement greater than 0.1\,m
in all three evaluated rollouts, excluding negligible-motion
segments for which displacement direction is unreliable.

As shown in Table~\ref{tab:progress_translation_analysis},
the with-progress checkpoint at step 1920 achieves a lower
mean physical displacement error than the without-progress
checkpoint at the same training step
(0.1736 versus 0.3523\,m), together with lower RMSE
(0.2009 versus 0.4153\,m).
Mean directional error also decreases from
$16.64^\circ$ to $11.01^\circ$.
The later without-progress checkpoint at step 2560
exhibits larger errors, reaching 0.6125\,m in mean
displacement error and $88.03^\circ$ in directional error.

The progress-loss checkpoint improves both translation magnitudes
and physical displacement estimates on the evaluated subset.
Each checkpoint uses independently generated rollouts with shared
episode identities; observations and executed trajectories can differ.

\paragraph{Connection to the Loss Formulation.}
The Directional Progress Loss penalizes insufficient projection
onto the GT translation direction for targets with non-negligible
translation:
\begin{equation}
    \mathcal{L}_{\mathrm{prog}}
    =
    \max\left(
        0,\,
        \rho d -
        \hat{\mathbf{t}}^{\top}
        \frac{\mathbf{t}^{*}}{d}
    \right),
    \qquad d=\|\mathbf{t}^{*}\|_2>\tau.
\end{equation}
Targets with $d\leq\tau$ are excluded, so the objective does not
encourage translation for negligible-motion targets.
We use $\rho=0.8$ and $\tau=0.1\,\mathrm{m}$.
Unlike a magnitude-only constraint, this term also accounts for
the predicted translation's alignment with the target direction.

Directional Progress Loss regularizes translation underestimation
in the adopted inverse-extrinsic parameterization; its direct
constraint acts on translation projection, not camera-center motion.
The reduced magnitude deficit and lower physical displacement errors,
together with the navigation ablation, support its effectiveness
under the evaluated conditions.

\section{Robustness to Controlled Visual Perturbations}
\label{app:visual_robustness}

We evaluate PoseVLN under two controlled visual perturbations:
brightness reduction by a factor of 0.5 and Gaussian blur with
radius 2. All other settings remain unchanged.

As shown in Table~\ref{tab:visual_robustness}, SR changes from
64.6\% on clean inputs to 62.5\% with brightness $\times 0.5$
and 58.4\% with Gaussian blur, decreases of 2.1 and 6.2 points.

Both conditions retain higher SR than the clean-input action-only
baseline (57.0\%). The larger degradation under blur indicates
greater sensitivity to lost visual detail. These experiments measure
robustness to the specified appearance changes.

\begin{table}[t]
\centering
\caption{
Navigation robustness under controlled visual perturbations
on R2R-CE Val-Unseen. Performance changes are measured relative
to the clean-input PoseVLN result.
}
\label{tab:visual_robustness}
\small
\setlength{\tabcolsep}{7pt}
\renewcommand{\arraystretch}{1.05}
\begin{tabular}{lcc}
\toprule
\textbf{Test Condition} &
\textbf{SR (\%) $\uparrow$} &
\textbf{$\Delta$SR (pp)} \\
\midrule
PoseVLN (Clean) & 64.6 & -- \\
Brightness $\times 0.5$ & 62.5 & $-2.1$ \\
Gaussian Blur ($r=2$) & 58.4 & $-6.2$ \\
\midrule
Action-only (Clean) & 57.0 & -- \\
\bottomrule
\end{tabular}
\end{table}

\section{Pose Context and Closed-Loop Collection Audit}
\label{app:pose_context_audit}

We distinguish pose context during optimization and DAgger
collection, reporting ground-truth and predicted pose inputs
and expert intervention statistics.

\paragraph{Pose Context During Optimization.}

During supervised optimization, all valid current and historical
pose inputs are constructed from simulator-recorded camera poses.
Therefore, 100\% of the valid pose context is ground-truth-derived,
while 0\% is model-predicted. The same recorded camera trajectories
are also used to construct the local $\Delta$Pose regression targets.

During policy-controlled DAgger rollouts, geometric states are
recursively propagated using predicted $\Delta$Pose, exposing
the policy to its own rollout distribution. Optimization continues
to use recorded pose annotations, as summarized in
Table~\ref{tab:pose_context_ratio}.

\begin{table}[t]
\centering
\caption{
Sources of current and historical pose context during
optimization. All valid pose inputs are constructed from
simulator-recorded camera poses.
}
\label{tab:pose_context_ratio}
\small
\setlength{\tabcolsep}{6pt}
\renewcommand{\arraystretch}{1.05}
\begin{tabular}{lcc}
\toprule
\textbf{Pose Source} &
\textbf{Current (\%)} &
\textbf{Historical (\%)} \\
\midrule
Ground-truth poses & 100 & 100 \\
Model-predicted poses & 0 & 0 \\
\bottomrule
\end{tabular}
\end{table}

\paragraph{Expert Intervention Frequency.}

During closed-loop collection, expert intervention is triggered
when the agent deviates from the reference path by more than
1.0~m, when the normalized accumulated action-error score exceeds
0.5, when the accumulated number of action errors exceeds 8,
or when the policy produces an invalid Stop action. The
path-recovery controller returns control to the policy when
the deviation falls below 0.5~m.

Table~\ref{tab:intervention_statistics} reports interventions
in 96.68\% of R2R episodes and 88.50\% of RxR episodes, with
1.09 and 1.01 mean takeovers per episode. Combined, the rate
is 91.32\% with 1.04 takeovers per episode. Takeover counts
exclude pose re-anchoring at policy return.

\begin{table}[t]
\centering
\caption{
Expert intervention statistics during closed-loop DAgger
collection. The intervention rate denotes the percentage
of completed episodes containing at least one expert takeover.
}
\label{tab:intervention_statistics}
\small
\setlength{\tabcolsep}{5pt}
\renewcommand{\arraystretch}{1.05}
\begin{tabular}{lcc}
\toprule
\textbf{Dataset} &
\textbf{Intervention Rate (\%)} &
\textbf{Mean Takeovers} \\
\midrule
R2R & 96.68 & 1.09 \\
RxR & 88.50 & 1.01 \\
Combined & 91.32 & 1.04 \\
\bottomrule
\end{tabular}
\end{table}

\paragraph{Action Conditioning and Pose Re-anchoring.}

The action decoder uses the shared multimodal representation,
while CameraHead uses the current pose-token hidden state.
The pose head predicts $\Delta$Pose jointly with actions and
is not explicitly conditioned on the final generated or executed
action sequence; it does not integrate decoded actions.

During expert--policy handoffs in DAgger collection, the internal
pose state is re-anchored to the simulator-recorded camera pose
in the original episode-relative coordinate frame. Unexecuted
action predictions and stale recurrent states are discarded
before policy control resumes. This prevents expert-executed
motion from leaving the propagated state inconsistent with
the subsequent visual observation.

Re-anchoring is limited to DAgger collection. During evaluation,
non-initial poses are recursively predicted without GT correction
or external odometry. The effects of predicted-context optimization
and removing collection-time re-anchoring remain to be evaluated.

\section{Statistical Reliability and Batch-One Inference Efficiency}
\label{app:statistical_efficiency}

We assess navigation reliability using episode-level bootstrap
analysis on R2R-CE and RxR-CE Val-Unseen, and inference efficiency
using batch-one end-to-end decision latency.

\paragraph{Episode-Level Statistical Reliability.}
On R2R-CE Val-Unseen, we compare StreamVLN and PoseVLN
using 20,000 paired bootstrap resamples of the same
1,839 episodes, preserving episode correspondence.

\begin{table}[t]
\centering
\caption{
Statistical reliability and batch-one inference efficiency
on R2R-CE Val-Unseen.
Confidence intervals are estimated using 20,000 paired
episode-level bootstrap resamples.
Latency measurements report mean median and mean P95
end-to-end decision times.
}
\label{tab:bootstrap_results}
% 如果去掉缩放后发现表格太宽，可以取消下面一行的注释并适当调小数值（默认是6pt）
% \setlength{\tabcolsep}{4pt}
\begin{tabular}{@{}lccc@{}}
\toprule
\textbf{Method}
& \textbf{SR (\%) [95\% CI]}
& \shortstack{\textbf{Mean Median}\\\textbf{Latency (ms)}}
& \shortstack{\textbf{Mean P95}\\\textbf{Latency (ms)}} \\
\midrule
StreamVLN
& 56.2 [53.94, 58.40]
& 546.97
& 767.83 \\
PoseVLN
& \textbf{64.6 [62.43, 66.78]}
& 560.12
& 781.09 \\
Difference
& +8.4 pp [5.87, 11.09]
& +13.15 (+2.40\%)
& +13.26 (+1.73\%) \\
\bottomrule
\end{tabular}
\end{table}

\paragraph{Statistical Reliability and Efficiency Results.}
Table~\ref{tab:bootstrap_results} shows an 8.4-point SR gain
over StreamVLN, with a paired 95\% confidence interval
of [5.87, 11.09] points, entirely above zero.

On RxR-CE English Val-Unseen, PoseVLN achieves
62.0\% SR with a 95\% confidence interval of
[60.45, 63.58] across 3,669 evaluation episodes.
These intervals quantify episode-level uncertainty
for the evaluated checkpoints.

The navigation improvement comes with only 2.40\%
additional mean median latency and 1.73\% additional
mean P95 latency, corresponding to increases of
13.15\,ms and 13.26\,ms, respectively.
Together, these results demonstrate a navigation gain
with modest inference overhead under the evaluated
batch-one setting.

\paragraph{Batch-One Inference Efficiency.}

Table~\ref{tab:inference_latency} summarizes batch-one
end-to-end latency overhead relative to StreamVLN:
2.40\% in mean median latency and 1.73\% in mean P95 latency.
These results quantify the added inference cost of pose encoding
and local pose prediction.

\begin{table}[t]
\centering
\caption{
Relative end-to-end inference latency overhead of PoseVLN
compared with StreamVLN under batch-one inference.
}
\label{tab:inference_latency}
\small
\setlength{\tabcolsep}{6pt}
\renewcommand{\arraystretch}{1.05}
\begin{tabular}{lc}
\toprule
\textbf{Latency Metric} &
\textbf{Relative Overhead (\%)} \\
\midrule
Mean median latency & +2.40 \\
Mean P95 latency & +1.73 \\
\bottomrule
\end{tabular}
\end{table}

\section{Fixed-Scale Position Error Comparison with VGGT}
\label{app:vggt_fixed_scale}

We complement the per-pair scale-aligned comparison in Table~4
with fixed-scale position errors over $H\in\{4,8,12,16\}$.
VGGT uses one global translation scale calibrated on R2R
training data, fixed across evaluation pairs and horizons.

\paragraph{Fixed-Scale Calibration Protocol.}
We estimate a single global translation scale using
103,300 moving H=4 image pairs from 10,819 reference
trajectories across 61 R2R training scenes.
We verified that all calibration episode IDs belong to
the R2R training split.
The scale is obtained by minimizing the translation error
over the calibration set $\mathcal{C}$:
\begin{equation}
    s^*
    =
    \underset{s>0}{\arg\min}
    \sum_{p\in\mathcal{C}}
    \left\|
        s\mathbf{t}^{\mathrm{VGGT}}_p
        -
        \mathbf{t}^{\mathrm{GT}}_p
    \right\|_2^2
    =
    1.7772.
\end{equation}
Here, $\mathbf{t}^{\mathrm{VGGT}}_p$ and
$\mathbf{t}^{\mathrm{GT}}_p$ denote the VGGT-predicted
and GT translation vectors, respectively.
The resulting scale is shared across all evaluated horizons
$H\in\{4,8,12,16\}$ and both R2R-CE and RxR-CE evaluations.
PoseVLN retains its original metric-scale predictions
without additional scaling.

Evaluation GT is used for position-error measurement and
moving-pair selection, never for scale fitting. Scale calibration
uses R2R training trajectories that are separate from the
evaluation episodes.

\paragraph{Fixed-Scale Position Error Results.}
As shown in Table~\ref{tab:vggt_fixed_scale}, the two methods
have comparable mean position error at $H=4$
(0.2568\,m for PoseVLN and 0.2547\,m for VGGT).
At longer horizons, PoseVLN achieves lower errors:
0.5049 versus 0.6772\,m at $H=8$,
0.7863 versus 1.0563\,m at $H=12$, and
1.1092 versus 1.4348\,m at $H=16$.
Both methods exhibit increasing error with horizon,
while PoseVLN shows lower mean position errors at
$H\in\{8,12,16\}$ under this fixed-scale evaluation.

\begin{table}[t]
    \centering
    \small
    \setlength{\tabcolsep}{9pt}
    \caption{
        Mean position error (m, lower is better) across prediction
        horizons without per-pair translation scale alignment.
        VGGT uses a single global translation scale calibrated
        on R2R training data and shared across all evaluated
        horizons and benchmarks.
    }
    \label{tab:vggt_fixed_scale}
    \begin{tabular}{@{}rrr@{}}
        \toprule
        \textbf{Horizon $H$}
        & \shortstack{\textbf{PoseVLN}\\\textbf{error (m)}}
        & \shortstack{\textbf{VGGT (fixed scale)}\\\textbf{error (m)}} \\
        \midrule
        4  & 0.2568 & \textbf{0.2547} \\
        8  & \textbf{0.5049} & 0.6772 \\
        12 & \textbf{0.7863} & 1.0563 \\
        16 & \textbf{1.1092} & 1.4348 \\
        \bottomrule
    \end{tabular}
\end{table}

\paragraph{Observation Protocols and Interpretation.}
PoseVLN predicts causally from intermediate RGB observations,
history, and propagated poses; VGGT uses endpoint image pairs.
These are distinct observation protocols. The fixed-scale results
provide complementary evidence
of PoseVLN's longer-horizon geometric accuracy without
per-pair scale alignment.

% IMPORTANT BEFORE SUBMISSION: State exactly how the fixed VGGT
% translation scale was obtained (value, calibration data, whether
% shared across horizons), and confirm that evaluation uses the same
% valid prediction--GT pairs and position-error definition as PoseVLN.
% Do not present the protocol as matched-observation comparison unless
% that experiment has actually been performed.

\section{Auxiliary Geometric Supervision Ablation on RxR-CE}
\label{app:rxr_aux_supervision}

We extend the ablation in Table~\ref{tab:loss_ablation}
to RxR-CE English Val-Unseen. Auxiliary-only retains action,
pose, and directional progress objectives without numerical
pose inputs, while full PoseVLN uses pose-conditioned memory
and recursive pose feedback.

Table~\ref{tab:rxr_aux_supervision} shows improvements from
auxiliary-only to full PoseVLN: SR 56.0\% to 62.0\%, SPL
51.4\% to 53.5\%, and NE 5.4 to 4.8\,m. Together with
the R2R-CE comparison in the updated
Table~\ref{tab:loss_ablation} (57.5\% versus 64.6\% SR),
these results establish benefits on both benchmarks.

\begin{table}[t]
    \centering
    \small
    \setlength{\tabcolsep}{6pt}
    \caption{
        Auxiliary geometric supervision versus full pose-conditioned
        navigation on RxR-CE English Val-Unseen. Both variants use
        pose-related training objectives; only full PoseVLN uses
        recursively predicted pose tokens as navigation context.
    }
    \label{tab:rxr_aux_supervision}
    \begin{tabular}{@{}lrrr@{}}
        \toprule
        \textbf{Model}
        & \textbf{SR (\%) $\uparrow$}
        & \textbf{SPL (\%) $\uparrow$}
        & \textbf{NE (m) $\downarrow$} \\
        \midrule
        Auxiliary-only (no pose input)
        & 56.0 & 51.4 & 5.4 \\
        Full PoseVLN (predicted pose)
        & \textbf{62.0} & \textbf{53.5} & \textbf{4.8} \\
        \bottomrule
    \end{tabular}
\end{table}

\section{Real-World Navigation Evaluation}
\label{app:real_world}

\paragraph{Evaluation Setup.}
We evaluate PoseVLN on Unitree Go2 in outdoor (Scene 1)
and indoor (Scene 2) environments containing doorways,
corridors, and obstacles. Each scene includes three tasks
covering room exit, turning, stopping, and room entry.
Three trials per task with varied starting positions yield
18 trials across six settings. Tasks 1 and 6 use the same
instruction in different scenes.

\paragraph{Quantitative Results.}
Table~\ref{tab:realworld_results} reports 11 successes in
18 trials (61.1\%), with at least two successful trials
in four of six task settings. Tasks 3 and 5 yield no successes
in three trials each, involving wall-target stopping and room
entry, respectively.

\begin{table}[t]
\centering
\caption{Real-world navigation results on Unitree Go2
across an outdoor scene (Scene 1) and an indoor scene
(Scene 2). Each task setting is evaluated three times
with slightly varied starting positions.
SR reports the number of successful trials out of three.
S/F indicates whether at least two of the three trials
succeed.}
\label{tab:realworld_results}

\begingroup
\setlength{\tabcolsep}{3pt}
\renewcommand{\arraystretch}{1.2}
\footnotesize

\begin{tabular}{
@{}c>{\raggedright\arraybackslash}p{0.70\linewidth}cc@{}
}
\toprule
\textbf{ID} & \textbf{Instruction}
& \textbf{SR} & \textbf{S/F} \\
\midrule

\multicolumn{4}{@{}l}{\textit{Scene 1 (Outdoor)}} \\
\addlinespace[3pt]

1 &
Walk straight toward the open doorway, pass through it,
and stop once you have completely left the room.
& 3/3 & S \\

\addlinespace[3pt]
2 &
Walk straight toward the open doorway and pass through it.
Once you have completely left the room, turn right and stop.
& 2/3 & S \\

\addlinespace[3pt]
3 &
Walk straight toward the open doorway, pass through it,
and stop in front of the wall directly opposite the doorway.
& 0/3 & F \\

\midrule
\multicolumn{4}{@{}l}{\textit{Scene 2 (Indoor)}} \\
\addlinespace[3pt]

4 &
Walk straight through the open doorway into the hallway,
turn left to face down the hallway after you have completely
left the room, and stop once you have completed the left turn.
& 3/3 & S \\

\addlinespace[3pt]
5 &
Walk straight through the open doorway ahead and to your right,
and stop once you have completely entered the room.
& 0/3 & F \\

\addlinespace[3pt]
6 &
Walk straight toward the open doorway, pass through it,
and stop once you have completely left the room.
& 3/3 & S \\

\midrule
\textbf{Total} & \textbf{Six task settings}
& \textbf{11/18} & \textbf{4/6} \\
\bottomrule
\end{tabular}

\endgroup
\end{table}

\begin{figure}[t]
    \centering
    \includegraphics[width=\linewidth]{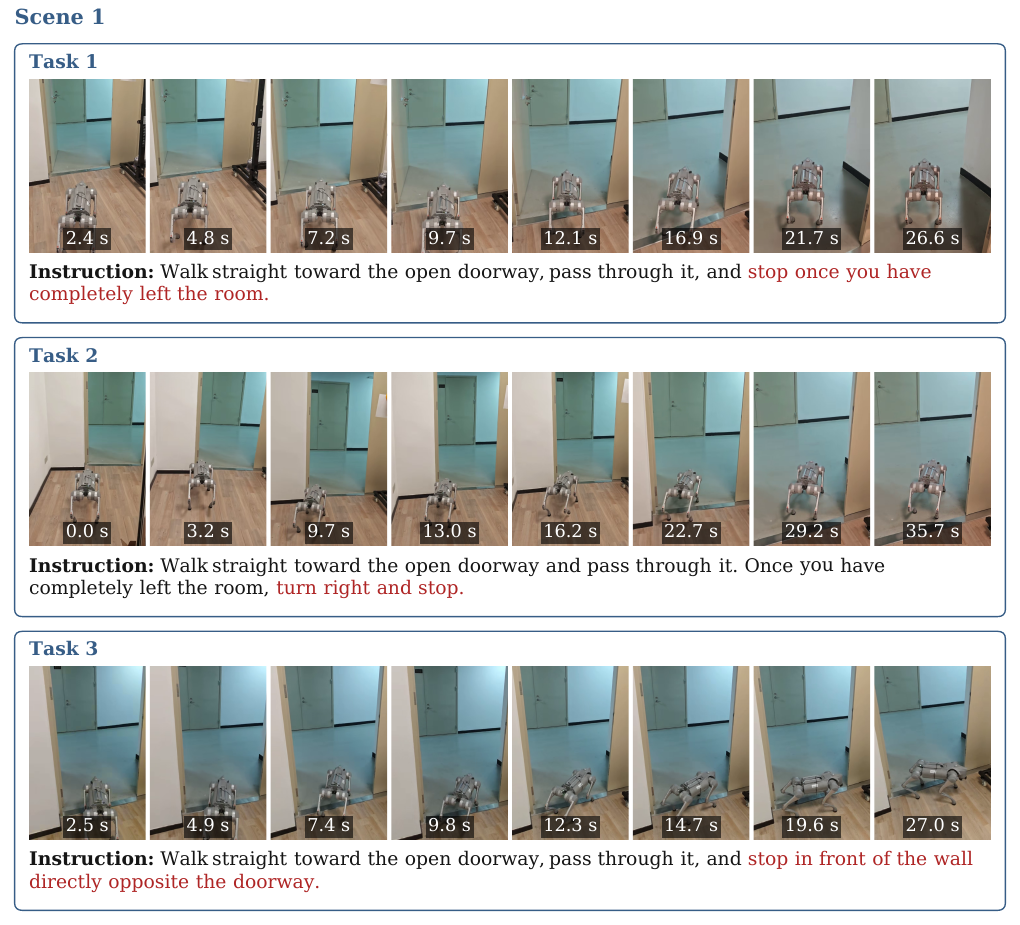}
    \caption{
    Representative real-world navigation trajectories
    of PoseVLN on Unitree Go2 in the outdoor scene (Scene 1).
    The three rows correspond to room exit (Task 1),
    room exit followed by a right turn (Task 2),
    and stopping at the wall opposite the doorway (Task 3).
    Frames are ordered chronologically from left to right,
    with source-video timestamps in seconds.
    The full instruction is shown below each trajectory.
    }
    \label{fig:realworld_scene1}
\end{figure}

\begin{figure}[t]
    \centering
    \includegraphics[width=\linewidth]{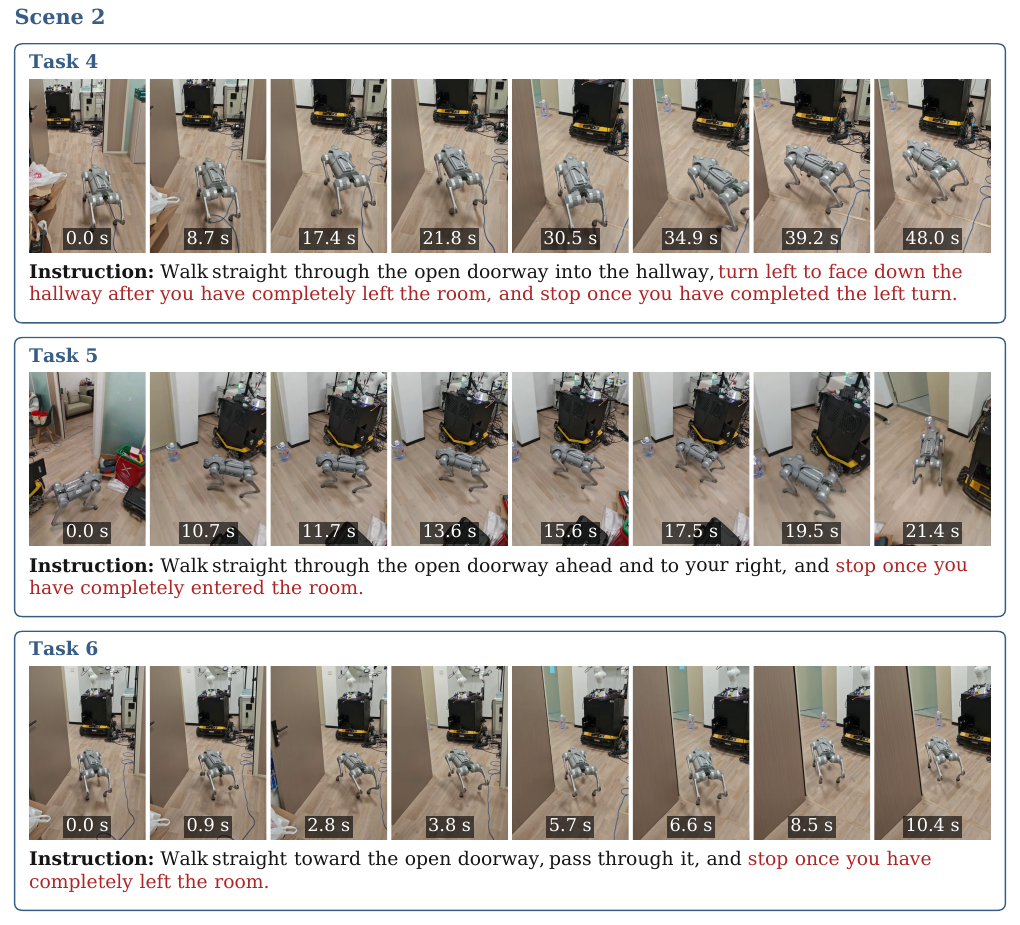}
    \caption{
    Representative real-world navigation trajectories
    of PoseVLN on Unitree Go2 in the indoor scene (Scene 2).
    The three rows correspond to room exit followed by
    a left turn (Task 4), entering the room through the
    doorway ahead and to the right (Task 5),
    and room exit (Task 6).
    Frames are ordered chronologically from left to right,
    with source-video timestamps in seconds.
    The full instruction is shown below each trajectory.
    }
    \label{fig:realworld_scene2}
\end{figure}

\paragraph{Trajectory Visualization.}
Figures~\ref{fig:realworld_scene1}
and~\ref{fig:realworld_scene2} visualize representative
recordings from the outdoor scene (Tasks 1--3)
and the indoor scene (Tasks 4--6), respectively.
Each figure contains three rows, with one recorded trial
per task.
Each row presents eight selected video frames in
chronological order from left to right, together with
the complete navigation instruction.
The timestamps indicate elapsed time in the source video;
the selected frames are not necessarily equally spaced.

The recordings show room exit, turning, targeted stopping,
and attempted room entry, including the unsuccessful settings
in Table~\ref{tab:realworld_results}.
The figures show individual trials; the table aggregates
all three repetitions per setting.

\paragraph{Discussion.}
The 11 successful trials (61.1\%) demonstrate RGB-based
language-guided navigation on Unitree Go2 in outdoor and indoor
settings, including doorway traversal and turning.
Tasks 3 and 5 remain challenging, with no successful trials
for wall-target stopping or specified room entry.
These results establish feasibility across the evaluated tasks
and motivate broader tests of reliability across environments
and instructions.

\section{Availability of R2R-CE Test Evaluation}
\label{app:test_evaluation}

The official VLN-CE test evaluation server was retired in January 2026,
and ground-truth test goals are not publicly available.
The organizers recommend using the Val-Unseen split for future method
comparisons\footnote{\url{https://eval.ai/web/challenges/challenge-page/719}}.
Following this recommendation, we report results on R2R-CE and RxR-CE
Val-Unseen, achieving 64.6\% and 62.0\% SR, respectively.

\end{document}